%% file: main.tex
\documentclass[11pt,fancychapters]{report}
\usepackage[a4paper, total={6in, 8in}]{geometry}
\usepackage{listings}
\usepackage{setspace}
\usepackage{hyperref}
\usepackage{acro}
\usepackage{amsmath}
\usepackage{amsthm}
\usepackage{graphicx}
\usepackage{geometry}
\usepackage{subcaption}
\usepackage{cancel}
\usepackage{tikz}
\usepackage{xcolor}
\usepackage{dsfont}
\usepackage{algorithmic}
\usepackage{algorithm}

\usepackage{natbib}

\definecolor{myred}{rgb}{0.93, 0.1, 0.1}
\definecolor{blue}{rgb}{0.2, 0.1, 0.9}
\definecolor{violet}{rgb}{0.54, 0.17, 0.89}

\input{./amacros}

\input{./math_commands.tex}

\title{\vspace{-5cm}{\normalsize {\color{myred} \textbf{This book draft is partial and contains only first two chapters. The full version of the book has been published and is now available on Web.  \\ ~ \\ {\color{teal}\underline{For citation, please use the following published version}: \\ ~ \\ Sahisnu Mazumder and Bing Liu. \textit{Lifelong and Continual \\Learning Dialogue Systems}. Springer Nature, 2024.}  \\~ \\ ~ \\}}}
	
{\color{blue}Draft Version}\\ ~\\~\\  \textbf{Lifelong and Continual Learning Dialogue Systems}}
\author{Sahisnu Mazumder \\ Bing Liu}
\date{November 11, 2022}

\begin{document}

\maketitle
\pagenumbering{gobble}
\newpage
\pagenumbering{roman}
\include{./bio}

\include{./fm}
\tableofcontents
\newpage
\pagenumbering{arabic}

\input{./ch01}
\input{./ch02}

\input{./ch03}

\input{./ch04}
\input{./ch05}
\input{./ch06}

\input{./ch07}
\input{./ch08}

\cleardoublepage
\bibliographystyle{abbrvnat}
\setcitestyle{authoryear,open={(},close={)}} 
\bibliography{book}

\end{document}

%% file: amacros.tex
\usepackage{subfiles}

\usepackage{enumitem}

\usepackage{amssymb}  

\usepackage{bbm}

\newcommand{\nb}{Na\"{i}ve Bayesian }

\newcommand{\data}{\mathcal{D}}
\newcommand{\task}{\mathcal{T}}
\newcommand{\E}{\mathbb{E}} 
\newcommand{\R}{\mathbb{R}} 
\AtBeginDocument{} 
	
\newcommand{\vectsmall}[1]%
	{{\mathbf{\MakeLowercase{\scriptstyle{#1}}}}} 

\newcommand{\transpose}{^{\rm T}} 

\let\emptyset\varnothing

\newcommand{\Lagr}{\mathcal{L}} 
\newcommand{\Lim}[1]%
	{\raisebox{0.5ex}{\scalebox{0.8}{$\displaystyle \lim_{#1}\;$}}} 

\makeatletter
\newcommand{\ALOOP}[1]{\ALC@it\algorithmicloop\ #1%
	\begin{ALC@loop}}
	\newcommand{\ENDALOOP}{\end{ALC@loop}\ALC@it\algorithmicendloop}

\makeatother

%% file: math_commands.tex
\usepackage{amsmath,amsfonts,bm}

\def\eqref#1{equation~\ref{#1}}

\def\1{\bm{1}}

\def\vx{{\bm{x}}}

\def\mX{{\bm{X}}}
\def\mY{{\bm{Y}}}

\DeclareMathAlphabet{\mathsfit}{\encodingdefault}{\sfdefault}{m}{sl}
\SetMathAlphabet{\mathsfit}{bold}{\encodingdefault}{\sfdefault}{bx}{n}



%% file: bio.tex
\chapter*{Authors' Biographies}
\markboth{AUTHORS' BIOGRAPHIES}{AUTHORS' BIOGRAPHIES}
\addcontentsline{toc}{chapter}{\protect\numberline{}{Authors' Biographies}}

\vspace{-0.3cm}
\section*{Sahisnu Mazumder}

\textbf{Web}: \url{https://sahisnu.github.io/} \\
\textbf{E-mail}: sahisnumazumder@gmail.com \\

\noindent
\small \textbf{Sahisnu Mazumder} is an AI Research Scientist at Intel Labs, Santa Clara, CA, USA where he works on Human-AI collaboration and dialogue \& interaction systems research.  He obtained his Ph.D. in Computer Science at the University of Illinois at Chicago (UIC), USA and his Masters in Computer Science from the Indian Institute of Technology (IIT) - Roorkee, India. His research interests include Lifelong and Continual Learning, Dialogue and Interactive Systems, Open-world AI / Learning, Knowledge base Reasoning, and Sentiment Analysis. He has published several research papers in leading AI, NLP and Dialogue conferences like AAAI, IJCAI, ACL, EMNLP, NAACL, SIGDIAL, CIKM; delivered tutorials in SIGIR-2022, IJCAI-2021 and Big Data Analytics (BDA - 2014) and served as PC Member / Reviewer of premier conferences like AAAI, IJCAI, ACL, EMNLP, NAACL, EACL, COLING and journals like ACM TALLIP and IEEE TNNLS. He has also worked as a Summer Research Intern at Huawei Research USA on projects related to user activity \& interest mining and at Microsoft Research - Redmond on Natural Language Interaction (NLI) system design. 

\normalsize

\section*{Bing Liu}

\textbf{Web}: \url{https://www.cs.uic.edu/~liub/} \\
\textbf{E-mail}: liub@uic.edu \\

\noindent
\small \textbf{Bing Liu} is a Distinguished Professor of Computer Science at the University of Illinois at Chicago. He received his Ph.D. in Artificial Intelligence from the University of Edinburgh. His current research interests include lifelong and continual learning, lifelong learning dialogue systems, open-world learning, sentiment analysis and opinion mining, machine learning and natural language processing. His previous interests also include fake review detection, Web mining and data mining. He has published extensively in top conferences and journals in these areas and authored four books: one about lifelong/continual machine learning, two about sentiment analysis, and one about Web mining. Three of his papers have received Test-of-Time awards and another one received Test-of-Time honorable mention. Some of his works have been widely reported in popular and technology press internationally. He served as the Chair of ACM SIGKDD from 2013-2017, as program chair of many leading data mining conferences including KDD, ICDM, CIKM, WSDM, SDM, and PAKDD, and as associate editor of many leading journals such as TKDE, TKDD, TWEB, and DMKD. He is the winner of 2018 ACM SIGKDD Innovation Award, and is a Fellow of AAAI, ACM, and IEEE. 

\normalsize

\clearpage

%% file: fm.tex
\newcommand{\sahisnu}[1]{{\color{red}{\small\bf\sf [sahisnu: #1]}}}
\newcommand{\bing}[1]{{\color{red}{\small\bf\sf [bing: #1]}}}


\chapter*{Abstract}
\noindent
Dialogue systems, commonly known as chatbots, have gained escalating popularity in recent times due to their wide-spread applications in carrying out chit-chat conversations with users and task-oriented dialogues to accomplish various user tasks. Existing chatbots are usually trained from \textit{pre-collected} and \textit{manually-labeled} data and/or written with \textit{handcrafted rules}. Many also use \textit{manually-compiled} knowledge bases (KBs).  Their ability to understand natural language is still limited, and they tend to produce many errors resulting in poor user satisfaction. Typically, they need to be constantly improved by engineers with more labeled data and more manually compiled knowledge.  This book introduces the new paradigm of \textit{lifelong learning dialogue systems} to endow chatbots the ability to learn continually by themselves through their own \textit{self-initiated} interactions with their users and working environments to improve themselves. As the systems chat more and more with users or learn more and more from external sources, they become more and more knowledgeable and better and better at conversing. The book presents the latest developments and techniques for building such continual learning dialogue systems that continuously learn new language expressions and lexical and factual knowledge during conversation from users and off conversation from external sources, acquire new training examples during conversation, and learn conversational skills. Apart from these general topics, existing works on continual learning of some specific aspects of dialogue systems are also surveyed. The book concludes with a discussion of open challenges for future research.

\vspace{7mm}
\textbf{Keywords:} \textit{Lifelong Machine Learning}; \textit{Lifelong Learning}; \textit{Continual Learning}; \textit{Dialogue Systems}; \textit{Interactive Systems}; \textit{Chatbots}; \textit{Conversational AI}; \textit{Virtual Assistants}; \textit{Learning during Conversation}; \textit{Learning after Model Deployment}; \textit{Open-world Learning}.

\vfill

\clearpage

\clearpage


{
\chapter*{Preface}
\addcontentsline{toc}{chapter}{\protect\numberline{}{Preface}}
\thispagestyle{plain}
\markboth{PREFACE}{PREFACE}

\noindent
The purpose in writing this book is to introduce the emerging topic of lifelong/continual learning dialogue systems. Dialogue systems, also commonly known as chatbots, are computer programs that can converse with humans to perform some intended tasks. They typically employ text or speech modes. As deep learning has improved accuracy of both speech recognition and text generation and processing dramatically since 2012 or so, dialogue systems are becoming increasingly ubiquitous and are used in all types of applications such as in smart phones, cars, home appliances, company web sites, mobile robots, etc. They provide a very rich set of services, e.g., performing some specific tasks and chit-chatting with human users. However, the user experiences have not been very satisfactory so far. It is clearly unfair to expect a deployed dialogue system to be able to understand everything that users may say, but these systems should be able to learn during conversation by interacting with users to become more and more knowledgeable and powerful. That is the goal of building lifelong learning dialogue systems. That is also the key motivation for us to write this book to introduce and to promote the research of lifelong learning dialogue systems.

The project of writing this book started with a tutorial titled ``\textit{Continual Learning Dialogue Systems - Learning on the Job after Model Deployment}'' that we gave at \textit{2021 International Joint Conference on Artificial Intelligence} (IJCAI-2021), August 21-26, 2021, Montreal, Canada. As we believe that lifelong learning dialogue systems is a very important topic for the future of dialogue systems and AI, we decided to develop the tutorial into a book. 
Our original interest in the topic stemmed from our research in lifelong/continual machine learning, dialogue systems, and natural language processing. Over the years, we have used many dialogue systems in smart phones and customer support websites. Our experiences have mostly been less than satisfactory. It is very natural to ask the question why the dialogue system cannot communicate with users and learn to improve itself when it cannot understand what the user says as we humans do. Most of the deployed dialogue systems work in multi-user environments, e.g., Amazon Alexa and Apple Siri, and Google Assistant. If these systems can learn even a tiny amount of knowledge from each user when it gets stuck, it will be very knowledgeable and smart over time. Another reason for our interest in lifelong learning dialogue systems is that our research group has been working on lifelong/continual learning for many years. This combination of factors encouraged us to work on the topic and to write this book. 

As lifelong learning dialogue systems sit at the intersection of dialogue systems in natural language processing and lifelong/continual learning in machine learning, this book will touch both fields. We aim to present a comprehensive survey and review of the important research results and latest ideas in these areas. We also want to propose a theoretical framework to be used to guide the future research and development in the field. This framework is called SOLA (\textit{Self-initiated Open-world continual Learning and Adaptation}), which is originally proposed for building autonomous and continual learning AI agents. Since lifelong learning dialogue systems are such agents, the framework is naturally suited for the topic. 
Presently, there are several research topics in dialogue systems that are closely related to lifelong learning dialogue systems. This book will bring all these topics under one roof and discuss their similarities and differences. Through this book, we would also like to motivate and encourage researchers to work on lifelong learning dialogue systems and practitioners to build lifelong learning dialogue systems that can be deployed for practical use to improve the user experiences and to make dialogue systems smart and smart over time. 
Without the capability of continually learning and accumulating knowledge, making inference about it, and using the knowledge to help future learning and problem solving, achieving true intelligence for AI agents is unlikely.

Two principles have guided the writing of this book. First, it should contain strong motivations for conducting research in lifelong learning dialogue systems in order to encourage graduate students and researchers to work on the problem. Second, the writing should be accessible to practitioners and upper-level undergraduate students who have basic knowledge of natural language processing and machine learning. Yet there should be sufficient in-depth materials for graduate students who plan to pursue Ph.D. degrees in dialogue systems, lifelong learning, or their integration of lifelong learning dialogue systems. We also strongly believe that lifelong learning dialogue systems can be built and deployed for practical applications.

This book is suitable for students, researchers, and practitioners who are interested in dialogue systems, natural language processing, and machine learning. Lecturers can readily use the book in class for courses in any of these related fields. 

\vspace*{2pc}
\noindent
Sahisnu Mazumder and Bing Liu\\
\noindent November 2022
}

 \clearpage



\chapter*{Acknowledgments}
\addcontentsline{toc}{chapter}{\protect\numberline{}{Acknowledgments}}
\thispagestyle{plain}
\markboth{ACKNOWLEDGMENTS}{ACKNOWLEDGMENTS}

\noindent
We would like to thank the current and former graduate students in our group and our collaborators: Joydeep Biswas, Jiahua Chen, Zhiyuan Chen, Daniel A. DeLaurentis, Sepideh Esmaeilpour, Neta Ezer, Geli Fei, Hasan Ghadialy, Yiduo Guo, Scott Grigsby, Estevam R. Hruschka Jr., Wenpeng Hu, Minlie Huang, Reid Hyland, Adam Kaufman, Zixuan Ke, Gyuhak Kim, Tatsuya Konishi, Mori Kurokawa, Huayi Li, Jian Li, Tianrui Li, Yanni Li, Haowei Lin, Zhou Lin, Lifeng Liu, Qian Liu, Guangyi Lv, Chuhe Mei, Piero Molino, Chihiro Ono, Arjun Mukherjee, Nianzu Ma, Shaoshuai Mou, Alexander Politowicz, Qi Qin, Steven Rizzi, Eric Robertson, Gokhan Tur, Vipin Vijayan, Lei Shu, Hao Wang, Mengyu Wang, Yijia Shao, Shuai Wang, Hu Xu, Yueshen Xu, Rui Yan, Yan Yang, Tim Yin, Tim Yuan, Philip S. Yu, Lei Zhang, Dongyan Zhao, Hao Zhou and Xiaoyan Zhu for their contributions of numerous research ideas and helpful discussions over the years. This book was also benefited significantly from numerous discussions in DARPA SAIL-ON Program meetings. 

\vspace{2mm}
Our greatest gratitude go to our own families. Sahisnu Mazumder would like to thank his parents Ramesh Ch. Mazumder and Dulali Mazumder, elder sister Snigdha, brother-in-law Sudip and (late) aunt Malaya Goswami for their invaluable support and encouragements. Bing Liu would like to thank his wife Yue, his children Shelley and Kate, and his parents. They have helped in so many ways.

\vspace{2mm}
The writing of this book was partially supported by a SAIL-ON Program Contract (HR001120C0023) of the Defense Advanced Research Projects Agency (DARPA), Northrop Grumman research gifts, three National Science Foundation (NSF) grants (1650900, IIS-1910424 and IIS-1838770), an NCI grant R01CA192240, and a Research Contract with KDDI. The content of the book is solely the responsibility of the authors and does not necessarily represent the official views of DARPA, Northrop Grumman, NSF, NCI, KDDI, UIC or Intel. 

\vspace{2mm}
The Department of Computer Science at the University of Illinois at Chicago provided computing resources and a very supportive environment for this project. We  would particularly like to thank Patricia Brianne Barrera, Sheri Lyn Joscelyn, Denise Marie Kelly, Emily Lam, Sherice Nelson, and Ivy Yuan for their valuable support in the research projects related to this book. 

\vspace{2mm}
Sahisnu Mazumder would also like to thank Intel Corporation, his colleagues at Multi-modal Dialogue and Interaction (MDI) group and Intelligent Systems Research (ISR) division at Intel Labs and specifically, Saurav Sahay and Lama Nachman for their invaluable support, advice and encouragement.

\vspace*{2pc}
\noindent Sahisnu Mazumder and Bing Liu\\
\noindent November 2022
 
\clearpage

%% file: ch01.tex
\chapter{Introduction}
\label{ch1}

Building \textit{dialogue systems} or \textit{conversational agents} 
capable of conversing with humans in natural language (NL) and understanding human NL instructions is a long-standing goal of AI \citep{winograd1972understanding}. These systems, also known as \textit{chatbots}, have become the front runner of AI advancement due to wide-spread applications such as assisting customers in buying products, booking tickets, reducing stress, and executing actions
like controlling house appliances and reporting weather information. However, the user experiences have not been fully satisfactory. There are many weaknesses with the current research and fielded dialogue systems. One of the major weaknesses is that they do not learn continuously during conversation (i.e., post-deployment) with the user after they are deployed in practice. Building lifelong learning dialogue systems that posses the capability of continuous learning during conversation is the main topic of discussion of this book. This chapter aims to motivate and provide the foundational idea of building such dialogue systems.  Note that, we use the term \textit{chatbots} to refer to all kinds of conversational agents, such as dialogue systems, personal assistants, conversational question-answering systems etc. onwards. 

In the following sections, we first provide some background of modern dialogue systems (Section \ref{ch1.sec1}) and discuss their general weaknesses (Section \ref{ch1.sec2}), which provide the motivations for studying and building lifelong learning dialogue systems (introduced in Sections \ref{ch1.sec3} and \ref{ch1.sec4}).  Finally, we conclude the chapter with a discussion on the organization of this book in Section \ref{ch1.sec5}.

\section{Dialogue and Interactive Systems: Background}
\label{ch1.sec1}
Dialogue systems  can broadly be categorized into two main types \citep{gao2019neural,jurafsky2020speech}: 

\begin{enumerate}
	\item[(1) ]  \textbf{Chit-chat systems}  \citep{vinyals2015neural,shang2015neural,yao2015attention,li2016deep,li2016diversity,serban2016building,chen2017survey,xing2017topic,wu2017sequential,shen2017conditional,mei2017coherent,serban2018survey,shen2018improving,wu2018neural,pandey2018exemplar} are chabots
	designed to engage users by conducting the chit-chat type of conversation on a wide range of topics \textit{without having a specific goal to complete.} Examples include \textit{Social Chatbots} like ELIZZA \citep{weizenbaum1966eliza}, PARRY \citep{colby1971artificial}, ALICE \citep{wallace2009anatomy},  Microsoft XiaoIce \citep{zhou2020design}, AliMe Chat \citep{qiu2017alime}, etc.  Such chatbots are built with the goal of supporting seamless conversation with users, and helping them with useful recommendations and mental supports.
	
	\item[(2)]  \textbf{Task-oriented chatbots} \citep{williams2007partially,bordes2016learning,wen2017network,budzianowski2018multiwoz,wen2017latent,lowe2017training,shah2018bootstrapping,zhao2017generative,luo2019learning} are chatbots designed to assist users to complete tasks based on users' requests, e.g., providing the requested information and taking actions.  Most of the popular personal assistants such as Alexa, Siri, Google Home, and Cortana, are task-oriented chatbots.  Besides, these types of chatbots are also built as QA Bots to support Question-answering (QA) over knowledge bases, conversational recommendation systems for online product or service recommendations to end-users and as Natural Language Interaction systems to enable natural language (NL) driven task completion. 
	
	Athrough the broad goal of task-oriented chatbots is to perform actions or tasks on users' behalf, they can also be of two types based on their design and the nature of interaction with users whom they are meant to support: \textbf{(1)} Systems that achieve task completion through \textit{multi-turn dialogues} with users where users express their intent and refine them (based on the feedback from the chatbot) over a sequence of dialogue turns with the system. They are formally known as \textbf{Task-oriented Dialogue Systems} (\textbf{ToDS}). \textbf{(2)} The other kind of systems that intend to accomplish tasks through a \textit{single-turn dialogue} where the user provides a NL instruction (command) and the system's goal is to just interpret it by translating it into some actions to be executed by the underlying application. Such systems are formally known as \textbf{Natural Language Interfaces} (\textbf{NLIs}). NLIs are also sometimes referred to as Natural Language Interaction systems in general. Although NLIs are mostly built to support a single-turn interaction with user per task completion goal, they can engage in muti-turn dialogues with user as well (similar to traditional ToDS systems) to better understand the NL instruction and resolving ambiguities (if any) to serve the user better.
\end{enumerate}

\begin{figure*}[t!]
	\vspace{0.3cm}
	\centering
	\includegraphics[height=4.1cm]{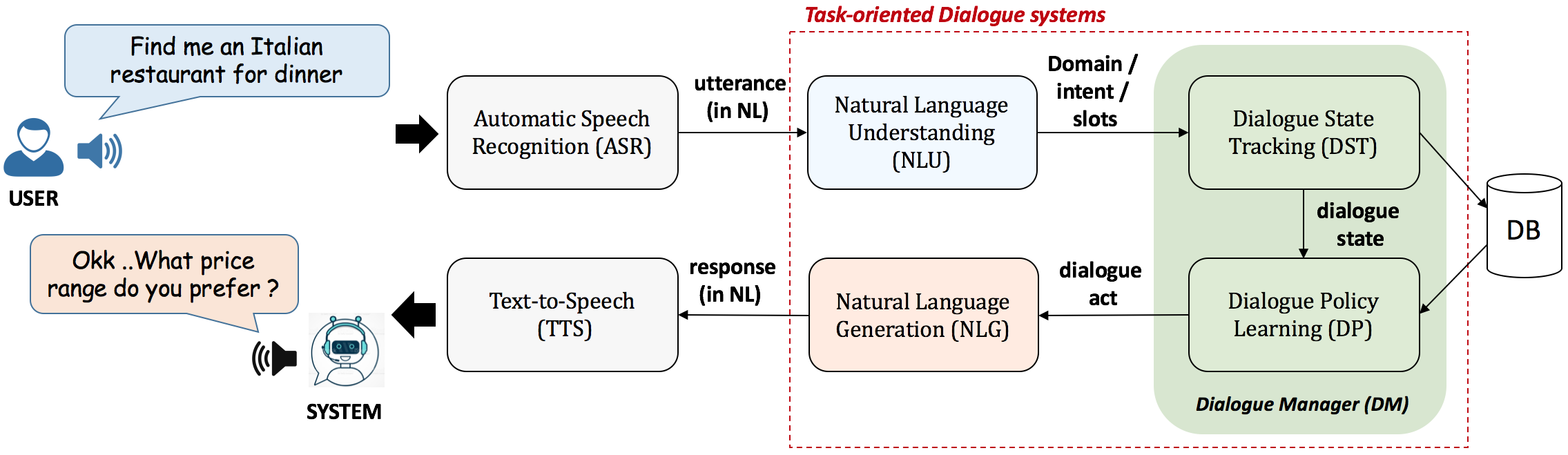}
	\caption{A typical task-oriented dialogue system with its modules.}
	\vspace{-0.2cm}
	\label{tods_modules}
\end{figure*}

\subsection{Task-oriented Dialogue Systems}

A full-fleged Task-oriented dialogue systems (ToDS) is mostly designed as a modular system, having six modules, viz., Automatic Speech Recognition (ASR), Natural Language Understanding (NLU), Dialogue State Tracking (DST), 
Dialogue Policy (DP) Learning, Natural Language Generation (NLG) and Text-to-speech (TTS) synthesis. Figure \ref{tods_modules} shows the architecture of a typical ToDS with all the modules integrated with each other. Here, the ASR module is responsible for translating the spoken utterance from the user into text which is fed to the NLU module for language understanding and the TTS module generates the speech from text which is the NL response generated by the NLG module. The DST and DP modules are often unified and referred to as Dialogue Manager (DM) which is responsible for progression of the dialogue by managing the dialogue turns. 

Often, ASR and TTS modules are studied as a seprate field of research and existing works mostly focus on the rest of four components and their interactions as a typical ToDS framework. In particular, they assume ASR and TTS are available to use and deal with only text-based user input and text-based generated ouput by the rest four modules. Thus, in the rest of our book, we mainly focus on NLU, DST, DP and NLG  as the four  main modules of a typical ToDS system.

\begin{enumerate}
	\item[(1)] \textbf{Natural Language Understanding (NLU):} The goal of the NLU module is to identify user intents and extract associated information (slots) from the user utterance. In general, NLU involves solving three subtasks:
	
	\begin{itemize}
		\item \textbf{\textit{Domain Classification}}, i.e., classifying the domain of the task expressed in user's utterance (e.g., is this user talking about airlines, movie, or music?)
		
		\item \textbf{\textit{Intent Classification}}, i.e., predicting the general task or goal that the user is trying to accomplish through the utterance (e.g., is this user wanting to search for a flight, book a movie ticket, or play a song,  etc.)
		
		\item \textbf{\textit{Slot filling}}, where the goal is to extract the particular slots and fillers that the user intends the system to understand from his/her utterance with respect to his/her intent. The problem of Slot filling is commonly formulated as a sequence lebelling task, where the goal is to tag each word of the utterance with one of the pre-defined semantic slot names. 
	\end{itemize}

	Let's have an example here. Considering the user utterance- ``\textit{Find me an italian restaurant for dinner}'', the NLU module will classify the domain as ``\textit{restaurant}" and intent as ``\textit{SearchRestaurant}" and extract slots as \{CUISINE: \textit{italian}, TIME: \textit{dinner} \}.

	\item[(2)] \textbf{Dialogue State Tracking (DST):} The DST module tracks the dialogue state that captures all the essential information in the conversation so far. Specifically the \textbf{dialogue state} includes the entire state of the frame at this point (the fillers of each slot), as well as the user’s most recent dialogue act, summarizing all of the user’s constraints. Below we show an example dialogue snippet, with the dialogue states tracked for each dialogue turn.
	
	\begin{quote}
		\texttt{User:}    Find me an italian restaurant for dinner. \\
		                           \hspace*{12mm} {\color{blue} \texttt{inform}(cuisine=\textit{Italian}; time=\textit{dinner})}  \\
		\texttt{System:}  Okk. What price range do you prefer? \\
		\texttt{User:}     may be in moderate  \\
		                           \hspace*{12mm} {\color{blue} \texttt{inform}(cuisine=\textit{Italian}; time=\textit{dinner}; price\_range=\$\$) }  \\
		\texttt{System:}  Sure. There an nice italian resturant at moderate price range nearby. \\
		\texttt{User:}     Awesome! What's the name and address? \\
		                           \hspace*{12mm} {\color{blue} \texttt{inform}(cuisine=\textit{Italian}; time=\textit{dinner}; price\_range=\$\$); \\  \hspace*{12mm}  request(name; address) } \\
		\texttt{System:}  The restaurant name is \textit{Franco's Ristorante }and address is   \\   
		                            \hspace*{14mm} \textit{300 W 31st St, Chicago, IL 60616}.
	\end{quote}
	
	\item[(3)] \textbf{Dialogue Policy (DP) Learning:} The DP Learning module selects the next action based on the current state (obtained from DST), i.e., deciding what action the system should take next, meaning what dialogue act to generate. That is, at a given dialogue turn $i$, DP predicts which action $\hat{A}_i \in A$ to take based on the entire dialogue state ($A_1, U_1, \dots, A_{i-1}, U_{i-1}, U_i$) [entire sequence of dialogue acts from the system ($A$) and from the user ($U$)]:
	\begin{gather}
		\hat{A}_i  = \arg \max_{A_i \in A} P(A_i | (A_1, U_1, \dots, A_{i-1}, U_{i-1}, U_i)) 
	\end{gather}
	{\color{black} For example, considering the first turn in the above dialogue snippet, with the captured dialogue state as \texttt{inform}(cuisine=\textit{Italian}; time=\textit{dinner}), the dialogue act generated by dialogue policy is \texttt{request}(price\_range). This results in the system's response where it asks the user for the price range in the dialogue turn.}
	
	\begin{quote}
    	\hspace*{12mm} {\color{blue} \texttt{inform}(cuisine=\textit{Italian}; time=\textit{dinner})} ~~~[ \textit{dialogue state} ] \\
    	\hspace*{50mm} $\Downarrow$\\
    	\hspace*{30mm} {\color{blue} \texttt{request}(price\_range)} ~~~[ \textit{dialogue act} ] 
    \end{quote}
	
	\item[(4)] \textbf{Natural Language Generation (NLG):} Once the policy has decided what speech act to generate (i.e., $\hat{A}_i$ ) for the current dialogue turn, the NLG module generates the text response to the user. It is often modeled in two stages, \textit{content planning} (i.e., what to say) and \textit{sentence realization} (i.e., how to say it). Sentence realization is commonly achieved through \textit{delexicalization} (i.e. mapping from frames to delexicalized sentences by often using encoder-decoder models). {\color{black} Here, we show an example corresponding to the last dialogue turn in the above dialogue snippet, where the system generates the delexicalized response for the given dialogue act  \texttt{inform}(name; address) as shown below and then, transforms it into an actual NL response by filling appropriate slots in the delexicalized response.}
	
	\begin{quote}
		\hspace*{25mm} {\color{blue} \texttt{inform}(name; address)} ~~~[ \textit{dialogue act} ]  \\
		\hspace*{50mm} $\Downarrow$  \\
		\hspace*{10mm} {\color{blue} The restaurant name is NAME\_SLOT and address is 
			 ADDRESS\_SLOT }  \\ \hspace*{35mm} ~~~~[ \textit{delexicalized response} ] 
	\end{quote}
\end{enumerate}

Although most of the existing ToDS systems are built as a modularized system, recent approaches have focused on training all these modules together end-to-end to reduce error propagation across modules.

\subsection{Chit-chat Systems}

Chit-chat Systems (or social chatbots) are often implemented as a unitary (non-modular) system. Based on their design approaches, they can broadly be  categorized into two types:

\begin{enumerate}
	\item[(1)] \textbf{Rule-based systems:} Rule-based systems work based on pattern and transform rules. Examples include ELIZA \citep{weizenbaum1966eliza} and PARRY \citep{colby1971artificial}. Each pattern/rule is linked to a keyword that might occur in a user utterance and trigers some transformation of a predfined template to generate response. For example, a pattern / transform rule in ELIZA is the following:
	
	\begin{quote}
		(0 YOU 0 ME)  ~~~~~~ \hfill{[pattern / decomposition rule]} \\
		\hspace*{10mm} $\Downarrow$ \\ 
		(WHAT MAKES YOU THINK I 3 YOU)   ~~~~~~~ \hfill{[transform / reassembly rule]}
	\end{quote}
   
   Here, ``0" in the pattern represents a kleen star (i.e., an indefinite number of words) and ``3" in the transform indicates that the third component of the subject decomposition is to be inserted in its place.  An example of this pattern / transform rule  applied in a dialogue turn would be- 
   
   \begin{quote}
    \texttt{User}:	You \textit{hate} me    \\
   	\texttt{System}: WHAT MAKES YOU THINK I \textit{HATE} YOU
   \end{quote}
	
	\item[(2)] \textbf{Corpus-based systems:} Corpus-based systems are trained to mimic human conversations by training on large amounts of human-human conversational data. Examples of this kind of system include modern chatbots like Microsoft XiaoIce \citep{zhou2020design}, AliMe Chat \citep{qiu2017alime} etc.
	
	Corpus-based systems are further categorized based on their response production method as:
	
	\begin{itemize}
		\item \textbf{\textit{Response by retrieval}} \citep{wang2013dataset,lowe2015ubuntu,wu2016topic,zhou2016multi,yan2016learning,wu2017sequential,bartl2017retrieval,chen2017survey,yang2018response,tao2019multi,tao2019one}: These methods formulate the response production task as an information retrieval (IR) problem. Considering the user’s turn as a query $q$, these methods aim to retrieve and repeat some appropriate turn $\hat{r}$ as the response from a corpus of conversations $D$ (training set for the system).  The turns in $D$ form a candidate set of responses, where each turn $r \in D$ is scored as a potential response with respect to the context $q$. The highest scored candiadate is then selected as response.
		\begin{gather}
			response(q, D) = \hat{r} = \arg \max_{r \in D} \frac{q . r}{|q| |r|}
		\end{gather}

        \begin{figure*}[t!]
        	\vspace{0.3cm}
        	\centering
        	\includegraphics[height=6cm]{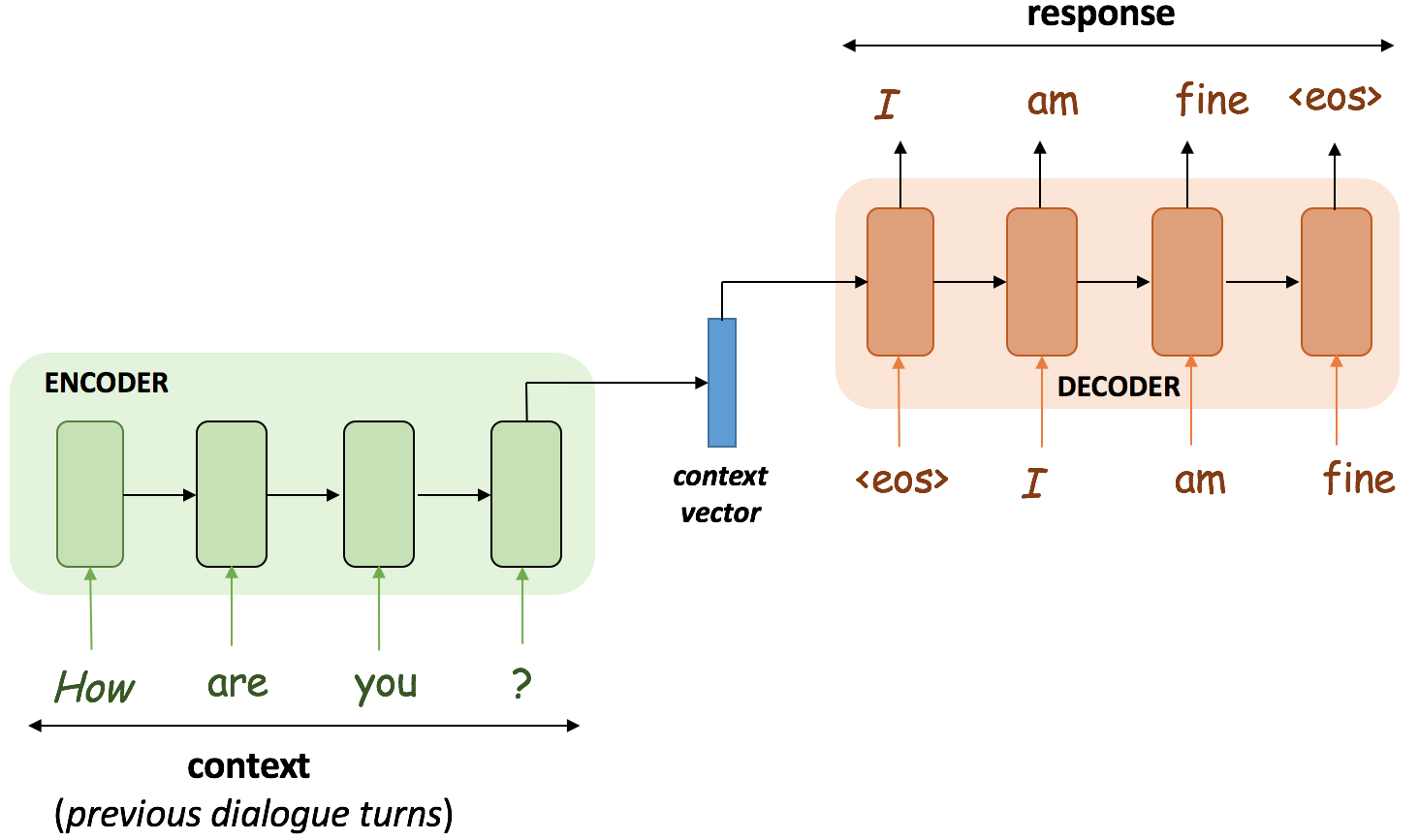}
        	\caption{Response generation by a typical seq2seq model.}
        	\vspace{-0.2cm}
        	\label{encoder_decoder}
        \end{figure*}
		
		\item \textbf{\textit{Response by Generation}} \citep{vinyals2015neural,shang2015neural,yao2015attention,li2016deep,li2016diversity,serban2016building,chen2017survey,xing2017topic,shen2017conditional,mei2017coherent,serban2018survey,shen2018improving,wu2018neural,pandey2018exemplar}: These methods view response production as an encoder-decoder task - transducing from the user’s prior turn to the system’s turn. 
		Figure \ref{encoder_decoder} shows an example of response generation by a typical Seq2seq model \citep{vinyals2015neural}, for a given conversation context. Here, the encoder is a Recurrent Neural Network (RNN) model that encodes the conversation context into a fixed-sized \textit{context vector} to summarize the information in the context and then, another RNN model takes the context vector as input and generates the response word-by-word.  
	\end{itemize}
\end{enumerate}

Although there have been a large number of works that solely focus on building a retrieval-based or generative chatbots as discussed above, there are approaches \citep{song2016two,serban2017deep,qiu2017alime} that adopt a hybrid strategy, i.e., combining neural generative and retrieval based models which has been shown to achieve better performance.

\section{Weaknesses of Modern Dialogue Systems}
\label{ch1.sec2}

Despite their widespread applications, chatbots still have some serious weaknesses:

\begin{enumerate}
	\item[(1)] \textbf{Dull response:} Existing social chatbots often generate uninformative responses like ``\textit{I don't know}", ``\textit{I don't have a clue}" etc.,  unspecific responses like ``\textit{I like all kinds of foods}" and repeated responses (i.e., telling the user same thing over and over again) due to lack of diversity \citep{li2016diversity,li2016deep,zhang2018generating,jiang2018sequence,see2019makes}.  Such dull or uninformative responses often disengage user from the conversation and leads to the abrupt end of the dialogue session. As we will discuss in Chapter~\ref{ch2}, such responses may also occur when the chatbot cannot detect novelty (what it does not understand) well and/or cannot characterize the novelty. As a result, it cannot give a suitable response. In fact, when such a situation occurs, it is opportunity for the chatbot to learn new knowledge.
	
	\item[(2)] \textbf{Out of context response:} Whenever the chatbot encounters some \textit{unknwon} topics or information in users utterance or \textit{changes in topics of discussion}, they often go out of context by generating off the track or irrelevant responses, dissatisfying the user. Like the case above, this is also the issue of not being able to detect novelty well and not being able to accommodate or adapt to the novelty, which we will discuss in Chapter~\ref{ch2}.
	
	\item[(3)] \textbf{Semantic Conflicts:} Open-domain chatbots often suffer from semantic conflict issues in conversation. Here is an example from \citep{huang2020challenges}:
	
	\begin{quote}
		\texttt{User:} I come from Shanghai. \\
		\texttt{System:} Cool! I am from Beijing, we come from the same city! \\
		\texttt{User:} Back to Beijing now. It is so cold but I am very happy! \\
		\texttt{System:} When will you come to Beijing?
	\end{quote}

   Here, although the user has responded that he/she is currently in Beijing, the chatbot asks when he intends to visit Beijing, creating semantic conflict in the conversation due to lack of understanding / reasoning over the context.
	
	\item[(4)] \textbf{Inconsistency:}  Both open-domain and task-oriented chatbots often suffer from inconsistency problem. Here is an example from \citep{huang2020challenges} showing the lack of consistency in open-domain chatbots due to the lack of a coherent personality:
	
	\begin{quote}
		\texttt{User:} Are you a boy or a girl? \\
		\texttt{System:} I am a boy. \\
		\texttt{User:} Are you a girl? \\
		\texttt{System:} Yes, I am a girl.
	\end{quote}
	
	\cite{schaub2021defining} also compiled a list of inconsistent behaviors often observed in modern task-oriented dialogue systems. These include misunderstanding of Intent, Bad entity, Repetition, “Bye” misunderstood, User insults, Bad API call, Incompleteness and Negation error.
	
	\item[(4)] \textbf{Interactiveness Issue:} Open-domain chatbots also suffer from bad interactiveness issue due to lack of grounding on the conversation context. Here is an example from \citep{huang2020challenges}, where the chatbot is not providing any recommendation inspite of being explicitly asked by the user and generating uninformative response.
	
	\begin{quote}
		\texttt{User:} Terrible breakfast today. \\
		\texttt{System:} How about we having breakfast together tomorrow morning? \\
		\texttt{User:} Sure. Would be great with your recommendation. \\
		\texttt{System:} Really?
	\end{quote}
	
	\item[(5)] \textbf{Lack of context-awareness:}  Due to the lack of context-rich datasets used in building chatbots, the deployed chatbots often lack context-awareness (e.g, understanding the situation of the user, dealing with ambiguous utterances from the user that needs reference resolution utilizing dialogue history, leveraging multi-model context inputs to generate responses etc.) while modeling their responses. This sometimes can lead to undesirable outcomes and dissatisfaction in users.
	
	\item[(6)] \textbf{Knowledge limitation due to fixed-corpus training:} A great deal of manual effort is needed to label training data or to write rules and compile knowledge bases (KBs). No matter how much data is used to train a chatbot, it is impossible to cover all possible variations of natural language. Thus, when deployed in practice, a well-trained chatbot often performs poorly. The pre-compiled KBs cannot cover the rich knowledge needed in practice. 
\end{enumerate}

A substantial amount of work has focused on solving the above weaknesses [1-5] over the years to come up with better response generation model that strives to produce more relevant and quality responses.  However, very little effort have been made to address the weakness-6 resolving which is the essential to build the next-generation dialogue and interactive systems. In this book, we focus on presenting and discussing methods that deal with weakness 6 which is the primary motivation for developing continual learning dialogue systems (as discussed next). Although here we primarily focus on weakness 6, solving this issue has an indirect influence on resolving other issues [1-5] as well, as this provides the chatbot the scope to learn continuously from users and self-improve its response generation capability over time as noted in the discussion of weaknesses 1 and 2.

\section{Motivation for Lifelong Learning Dialogue Systems}
\label{ch1.sec3}
As we can see, current dialogue systems still have many limitations. Most of us should have personal experiences of using dialogue systems either in our smart phones or on company websites that provide customer supports and the frustrations that we have experienced, and wished that real humans were behind such systems to serve us. Many researchers (including our own research group) have tried to address some of these issues in both task-oriented and chit-chat types of dialogue systems~\citep{zhou2018emotional,shu2019modeling,shu2019flexibly,hu2019gsn}. In this book, we focus on only the lifelong or continual learning aspect. In fact, several limitations can be addressed by continual/lifelong learning. 

The need for lifelong learning dialogue systems is a no-brainer simply because that is what we humans do and the current dialogue systems cannot do. Regardless of whether an existing dialogue system is based on hand-crafted rules or based on training via a deep learning model, it is an isolated system and can only be improved by human engineers either through adding more labeled training data or hand-crafted rules, or through improving the underlying deep learning models. After any such change, retraining of the whole system is typically required to update the system. This process is highly time consuming as training a deep learning model of a large system usually take days and weeks. However, this is not what we humans do. We humans have this remarkable ability to learn incrementally and quickly and we can learn by ourselves based on our own initiatives through our interactions with those who converse with us and the real-life environment, and learn from books and web pages without being supervised or guided by others. A dialogue system should do the same in order to achieve some level of true intelligence. 

Let us use various lifelong/continual learning settings to further motivate why dialogue systems should learn continually. Continual learning in machine learning is defined as learning a sequence of tasks incrementally~\citep{chen2018lifelong}. In supervised learning, each task is basically a classification problem that consists of a set of classes to be learned based on its training data. In dialogue systems, it can be a set of new skills that needs to be learned, e.g., booking a hotel and reserving a restaurant table. Continual learning can be further subdivided into two major settings, \textit{class continual learning} (CCL) (also called \textit{class incremental learning}) and \textit{task continual learning} (TCL) (also called \textit{task incremental learning}). In CCL, the classes in each task are distinct and no overlapping with classes of other tasks. In testing, no task related information is provided. During testing or application, the user will not identify which task/service he/she wants to use and the system has to automatically detect the task. For example, the user may say ``can you find a near-by Sushi restaurant for me.'' The system needs to first predict whether the user wants to find a restaurant or a hotel and then find some near-by Sushi restaurants and give to the user. To be able to incrementally add new skills is clearly very important for a dialogue system because retraining the whole system from scratch whenever a new skill is added is highly time consuming. In some cases, this kind of retraining may not be possible. For example, many companies do not have in-house capabilities to build a dialogue system for themselves and they licence systems from others. However, in almost all cases, they cannot change the systems in any way. If they want to add a new skill to provide some new services to their customers, they have go to the company from whom they licensed the dialogue system to request the addition. If the dialogue system has the ability to incrementally or continually learning, this problem is eliminated. The company can simply add the new skill by continual or incremental training using the data for the new skill. 

TCL differs from CCL in that TCL allows the user to indicate the task that he/she intends to perform. For example, he/she can select the option of using the restaurant subsystem. In TCL, each task model is usually built separately, but it can share knowledge with models of the previous tasks for knowledge transfer. For example, booking a hotel room and reserving a table in a restaurant have a lot of similarities and knowing one can help do the other. If the system has learned to book a hotel for the user before, it will be much easier for it to learn to help the user reserve a restaurant table, i.e., requiring much less training data. This kind of transfer in continual learning is automatic because the system decides from which previous tasks to transfer and what knowledge to be transferred~\citep{ke2020continual,ke2021achieving}. Clearly, this ability to transfer knowledge is highly desirable in building practical dialogue systems because manually labeling/writing a large amount of training dialogues is very costly and highly time-consuming. It is also important to note that this kind of transfer is different from the traditional transfer learning. Traditional transfer learning transfers knowledge from a selected source task to a selected target task and the selection is usually done by humans to ensure that the two tasks are similar and can achieve positive transfer. However, in continual learning, everything is done automatically with no manual involvement.

These two forms of learning based on existing research in continual learning, however, are still offline types of learning, meaning that both the tasks and their training data are provided by human engineers, and learning is done offline rather than on the fly during conversation with users. To achieve even more intelligence, it is important that the system can learn actively during conversation with users based on the initiation of the system itself in its interaction with users, i.e., learning on the job after deployment, from users. After all, we humans learn a great deal from our daily conversations.  

Learning during conversation or while working is highly desirable because of several reasons as a large number of unsatisfactory experiences or frustrations with current dialogue systems come from the fact that these systems cannot understand what the user is saying and give nonsense responses. The question is why the dialogue system cannot communicate with the user and learn to improve itself when it cannot understand a user utterance. If this type of learning is possible, then the user will not be so frustrated and the system will not have the same problem next time. Furthermore, most of the current dialogue systems work in multi-user environments, e.g., Amazon Alexa, Apple Siri, and Google Assistant. If these systems can learn even a tiny amount of knowledge from each user, the accumulated knowledge over time will make the system very smart and powerful. It is understandable that many users may not be willing to help the dialogue system learn but there should be many users who are willing to help. We will discuss this topic in greater detail in the next section. 


Another important aspect that needs continual learning is the learning of  personal traits, habits, temperaments,  emotional characteristics, and each user's personal situations so that the dialogue systems can be made more personalized to suit each individual user. This is especially important for dialogue systems that serve as personal assistants. To learn these types of information, the system needs to learn incrementally through a long history of dialogues. To improve this kind of learning and to learn quickly, the system can borrow knowledge from like-minded users and users with similar personal circumstances. In such cases, continual monitoring and incremental learning and automated knowledge transfer are all critical.  


The ultimate goal is for dialogue systems to achieve human-like behaviors and abilities in conversation so that they can continually learn and improve themselves on their own to become more and more knowledgeable and powerful without intervention from human engineers. Continual learning can be carried out either through interactions with users during conversation and/or from other sources such as books and web pages. Achieving this ultimate goal is clearly difficult at the moment, but it is possible to progressively improve the dialogue systems technology to move towards the goal gradually. 

\section{Lifelong Interactive Learning in Conversation}
\label{ch1.sec4}

In this section, we dive more into online continual learning during conversation rather than offline continual learning. This is perhaps the most central capability that a lifelong learning dialogue system should have. That is, a dialogue system should not be limited by offline training or pre-compiled knowledge bases (KBs). It should learn online on the fly during conversation continually, which is also called \textit{learning on the job or while working}. The process typically involves interaction with human users and learn to improve themselves in a self-motivated and self-initiated manner~\citep{chen2018lifelong,liu2020learning}. We humans also learn a great deal of our knowledge in our conversations and interactions with others which improve our knowledge and conversational abilities. We call this form of learning LINC (\textbf{\underline{L}ifelong \underline{IN}teractive learning in \underline{C}onversation})~\citep{mazumder2021job,liu2021lifelong}. LINC focuses on three continuous learning capabilities of chatbots: (1) learning factual knowledge in open-ended and information-seeking conversations, 
(2) learning to ground new natural language utterances, and (3) learning new conversational skills. Some initial attempts have been made in~\citep{mazumder2020continuous,mazumder2020application,mazumder2019lifelong,hancock2019learning,luo2019learning,mazumder2018towards}. 
A key idea for solving the LINC problem is to exploit \textbf{the wisdom of the crowd} in a multi-user environment (where almost all chatbots operate) to learn new knowledge by actively asking or interacting with the current user and/or other users to enable the chatbot to learn a large amount of knowledge quickly and effectively. In Chapter~\ref{ch2}, we will define a more general framework (called SOLA) for on-the-job learning. 

\textbf{LINC} can be more formally described as follows. During a conversation, the chatbot creates a new task $T_{N+1}$ on the fly when it wants to learn a piece of knowledge from a user utterance (e.g., extracting an unknown fact), encounters a problem (e.g., unable to understand a user utterance), encounters an unknown intent or slot in task-oriented dialogues, or is unable to answer a user query due to unknown entities or facts etc).\footnote{The knowledge learning tasks created by the chatbot for itself to learn are not the same as the tasks that the end-user wants to perform via the chatbot.} The problem of discovering the unknowns during conversation to formulate a new learning task is referred to as \textit{novelty detection} or \textit{open-set recognition} \citep{liu2020learning}, solving which is essential for LINC. Once the new task $T_{N+1}$ is formulated, the chatbot needs to acquire the ground truth training data $D_{N+1}$ for learning. To do this, the chatbot needs to \textit{formulate a dynamic interaction strategy} $\mathcal{S}$ to interact with the user, i.e., to decide what to ask and when to ask the user, and then \textit{execute} $\mathcal{S}$ to acquire the ground truth data. It then incrementally learns task $T_{N+1}$ with only one or a few examples. 

Existing approaches to obtaining the training data is through manual labeling or writing, which is both costly and time-consuming. As chatbots typically work in multi-user environments, we can exploit such environments to obtain the ground truth training data interactively during actual conversations. This process incurs no cost. 

Apart from traditional supervised learning through training using labeled examples, a dialogue system can learn during conversation in many other ways. 

\textbf{1. Extracting data or information directly from user utterances} 
(or dialogue history), which can be real-world facts, user preferences, etc. For example, while conversing about movies, if the user says ``\textit{I watched Forest Gump yesterday. The movie was awesome. Liked Tom Hanks' performance very much.}", the chatbot can extract the new fact (\textit{Forest Gump, isa, movie}) and (\textit{Tom Hanks, performed\_in, Forest Gump})~\citep{liu2020lifelong}. Later, the chatbot can use these facts in future conversations while answering questions like ``\textit{Who acted in Forest Gump?}" or generating a response to user's utterance ``\textit{I'm feeling bored. Can you recommend a good movie?}". 
The chatbot may even ask the user some related questions~\citep{liu2020lifelong} to obtain more knowledge. For example, after obtaining (\textit{Forest Gump, isa, movie}), the chatbot may ask a property question: ``\textit{What is the genre of Forest Gump?}" If the user answers, then another piece of knowledge is learned. Note that the extraction method proposed in~\citep{liu2020lifelong} is rule-based, which works with rule-based chatbots. Many deployed chatbots in industry are written with handcrafted rules.  
Additional knowledge/data may be inferred from the acquired knowledge and the existing  knowledge bases.


\textbf{2. Asking the current user} when the chatbot (1) doesn't understand a user utterance, or (2) cannot answer a user query, which forms a new learning task. To obtain the ground truth data, for (1), the agent may ask the current user for clarification, rephrasing, or even demonstration if it is supported~\citep{mazumder2020application}. For (2), it may ask the user for some supporting facts and then infer the query answer~\citep{mazumder2019lifelong,mazumder2020continuous}. 


\textbf{3. Asking other users} to obtain the answers when the chatbot could not answer a user query. For example, if a user asks  ``\textit{What is the capital city of the US?}" and the chatbot is unable to answer or infer now, it can try to find a good opportunity to ask another user later ``\textit{Hey, do you happen to know what the capital city of the US is?}" If the user gives the answer ``\textit{it's Washington DC}," the chatbot acquires the ground truth (a piece of new knowledge) which can be learned and used in future conversations. Note that although the answer cannot help the user who asked the question originally, it may be used in the future when a similar question is asked by another user. 

\textbf{4. Observing user demonstrations}. In some cases, the chatbots deployed in practice also come with Graphical User Interfaces (GUIs) or remote control facilities to explicitly control devices apart from controlling them via natural language commands. Examples of such systems include robots performing household tasks like cleaning robots and personal assistants integrated with home appliances like Smart TVs, Smart Lights, Smart Speakers, etc. 
Considering the user has issued an command and the bot has failed to execute the intended action, the user may perform the intended action via the GUI or remote control. The bot can record the sequence of executed action(s) performed by the user by accessing the underlying application logs and store the executed APIs as ground truth for the input natural language command. The command along with the invoked APIs can serve as labeled examples for learning the command \citep{forbes2015robot,wang2017naturalizing}. 

\textbf{5. Extract ground-truth data from external sources}, e.g., online documents or online knowledge bases~\citep{mitchell2018never}.

Acquiring knowledge from end-users comes with a shortcoming. That is, the knowledge learned from them can be erroneous. Some users may even purposely fool the system by providing wrong information or knowledge. 
Since chatbots usually work in a multi-user environment, such issues can be addressed through \textit{\textbf{cross-verification}}. After acquiring a piece of new knowledge (a new command pattern or a fact) in an interaction session, the agent can store these new examples in a unverified knowledge buffer. Next, while interacting with some other users in future sessions to accomplish a related task, it can ask these users to verify the accumulated unverified knowledge. Once verified for $K$ times (by $K$ different random users), the knowledge can be considered as trustworthy and removed from the unverified buffer and used in learning or chatting.

\section{Organization of the Book}
\label{ch1.sec5}

This book surveys and introduces the topic of lifelong or continual learning dialogue systems. Although the body of literature is not particularly large, many related papers are published in a number of conferences and journals. There is also a large number of papers that do not exhibit all the characteristics of a continual learning dialogue system and are somewhat \textit{weakly-related} to the topic. It is thus hard, if not impossible, to cover all of the important and related  works in the field. As a result, this book should not be taken as an exhaustive account of everything on this topic. However, we believe, the book provides a fairly broad coverage and presents some of the representative works and sets the necessary foundation for future advancements in this area. {\color{black} Also, this book mainly focuses on the advancement in the topic of continual learning dialogue systems. It does not intend to cover the basics of prerequisite knowledge (e.g., machine learning, dialogue systems, NLP, etc.) in depth. Interested readers can follow some of the closely-related books \citep{mctear2020conversational,jurafsky2020speech,gao2019neural,jokinen2009spoken,goodfellow2016deep,chen2018lifelong} on these topics for prerequisites.}

The rest of book is organized as follows. In Chapter~\ref{ch2}, we introduce the framework for open-world continual learning, namely, \textit{Self-initiated Open-world continual Learning and Adaptation} (SOLA), which sets the foundational idea for building a continual learning dialogue system.  

In Chapter~\ref{ch3}, we discuss various opportunities for \textit{continuous factual knowledge learning in dialogues}, and discuss methods about how a chatbot can learn by extracting factual knowledge from conversation, acquire lexical knowledge and facts interactively from end-users and learning by extracting knowledge from the Web.
 
Chapter~\ref{ch4} discusses and present methods for \textit{continual and interactive language learning} for natural language interface (NLI) design. In particular, we discuss about interactive language learning in games, building self-adaptive NLIs by continuously learning new commands from users and interactive semantic parsing for knowledge base question answering with user feedback to deal with parsing errors and ambiguities.

Chapter~\ref{ch5} is focused on \textit{continual learning in chit-chat dialogue systems}. Here, we discuss approaches to predict user satisfaction in conversation and present methods that use user dissatisfaction as an implicit signal to acquire user feedback and leverage it to continuously improve response generation after model deployment. 

In Chapter~\ref{ch6}, we discuss methods for building \textit{continual learning task-oriented dialogue systems} (ToDS). The chapter discusses the topic of open-world intent learning and present recent continual learning approaches on various ToDS sub-tasks like slot filling, dialogue state tracking, natural language generation, and also, an approach that attempts to jointly solve all these sub-tasks in an end-to-end continual learning setting.

Chapter~\ref{ch7} provides the motivation for \textit{continual learning of various  conversational skills} like personalized conversation modelling, learning of emotions, moods, and opinions in conversation and situation-aware conversation modeling. To the best of our knowledge, we have not come across any work in this topic that is particularly related to continual learning. Thus, here we provide a brief survey of existing works and describe an outline for future research in the topic. 

Finally, Chapter~\ref{ch8} concludes the book and discusses some challenges and future directions of research.


%% file: ch02.tex
\chapter{Open-world Continual Learning: A Framework}
\label{ch2}

As more and more AI agents are used in practice, we need to think about how to make these agents fully autonomous so that they can \textbf{(1)} learn by themselves continually in a \textit{self-motivated} and \textit{self-initiated} manner rather than being retrained offline periodically on the initiation of human engineers and \textbf{(2)} accommodate or adapt to unexpected or novel circumstances. As the real-world is an open environment that is full of unknowns or novelties, \textit{detecting} novelties, \textit{characterizing} them, \textit{accommodating or adapting} to them, and \textit{gathering} ground-truth training data and \textit{incrementally learning} the unknowns/novelties are critical to making AI agents more and more knowledgeable and powerful over time. The key challenge is how to automate the process so that it is carried out continually on the agent's \textit{own initiative} and \textit{through its own interactions} with humans, other agents and the environment just like human \textit{on-the-job learning}. 

This chapter develops a theoretical framework for open-world continual learning, which also serves as a framework for lifelong learning dialogue systems because such a dialogue system works in an open environment. Section~\ref{sec.cml} will briefly describe a dialogue system that follows the proposed framework. Since we want the framework to be applicable to AI agents in general, the framework will cover aspects that may not be necessary for a lifelong learning dialogue system. The key aspect of the framework is \textit{self-initiation}, which involves no engineers. The proposed framework is called 
\textit{\textbf{S}elf-initiated \textbf{O}pen-world continual \textbf{L}earning and \textbf{A}daptation} (\textbf{SOLA)}. SOLA is like human \textit{learning on the job} or \textit{learning while working}. It learns {after model deployment}. 

We should note that the SOLA framework goes beyond the traditional concept of machine learning or continual/lifelong learning, which normally starts with a labeled training dataset given to a machine learning algorithm to produce a model (e.g., a classifier). Thus, existing paradigms of \textit{classical machine learning} and \textit{lifelong}/\textit{continual learning} are only two aspects of the SOLA paradigm because in order to learn on the job, the system must discover and create new tasks to learn and also acquire labeled ground-truth training data during application on the fly by the agent itself. Furthermore, the agent must adapt to or accommodate unknowns or novelties. Before discussing the SOLA framework, let us first discuss the classical machine learning paradigm.  

\section{Classical Machine Learning}
\label{ch2.sec.classic}
The current dominant paradigm for machine learning (ML) is to run an ML algorithm on a given dataset to generate a model.
The model is then applied on a real-life performance task.
This is true for both supervised learning and unsupervised learning.
We call this paradigm {\em closed-world learning} because it makes the  \textit{independent and identically distributed} (I.I.D) assumption and it does not consider any other related information or the previously learned knowledge. Intuitively, by closed-world learning~\citep{fei2016breaking,liu2020learning,fei2016learning}, 
we mean what the agent sees in testing or application have been seen in training (we will discuss this further later). 

\begin{figure*}[ht!]
\centering
  \includegraphics[scale=0.80]{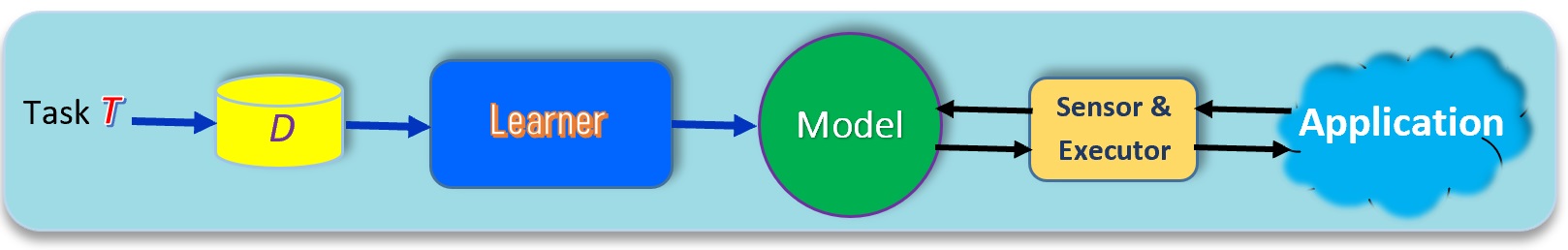}
  \caption{Architecture of the classical machine learning paradigm, where $T$ is the task and $D$ is its training data. The links in blue reflect the learning process and the links in black reflect the application process of the learned model. 
}
  \label{fig:traditional-ML}
\end{figure*}

Figure~\ref{fig:traditional-ML} illustrates the \textit{classical isolated learning paradigm}, which consists of two stages: (1) \textit{model building} and (2) \textit{model application} or deployment. In model building, the training data $D$ of task $T$ is used by the \textbf{Learner} (a ML algorithm) to produce a \textbf{Model}. This process is reflected by the blue links in the figure. In model application or deployment, the input data sensed by the \textbf{Sensor} from the application is sent to the \textbf{Model}, which produces a decision or action to be executed by the \textbf{Executor} in the application environment. This process is reflected by the black links. This classical learning paradigm has several limitations.

(1). Most real-life learning environments do not satisfy the I.I.D assumption because they are often dynamic and open, meaning that there may be constant data distribution changes and unknown objects appearing. In such cases, periodical retraining is constantly needed typically initiated by human engineers. To do retraining, manual labeling of new or additional training data is required, which is very labor-intensive and time-consuming.
Since the world is too complex and constantly changing, labeling and retraining need to be done periodically. This is a daunting task for human annotators and engineers. This is also the case for building and maintaining a chatbot. Even for unsupervised learning, collecting a large volume of data constantly may not be possible in many cases.

(2). It learns in isolation, meaning that it does not retain or accumulate knowledge learned in the past and use it to help future learning to achieve knowledge transfer. This is in sharp contrast to our human learning.
We humans never learn in isolation. We accumulate and maintain the knowledge learned from previous tasks and use it seamlessly in learning new tasks and solving new problems. That is why whenever we encounter a new situation or problem, we may notice that many aspects of it are not really new because we have seen them in the past in some other contexts.
When faced with a new problem or a new environment, we can adapt our past knowledge to deal with the new situation and also learn from it. Over time we learn more and more. and become more and more knowledgeable and more and more effective at learning. In the context of chatbots, it is easy to see opportunities for knowledge transfer. For example, after learning to book a meeting room, learning to reserve a table in a restaurant should be easier. 

(3). There is no learning on the job after the model has been deployed (also called post-deployment). Human learning is different as we continue to learn on the job after formal training. Studies have shown that about 70\% of human knowledge is learned
while working on a task or on the job. Only about 10\%
is learned through formal training and the rest 20\% is learned through imitation of others. An AI system should also learn on the job during model applications. In the context of chatbots, this means to learn during conversation with users. 

In the next section, we use a motivating example to illustrate that this classical paradigm is not sufficient in many practical applications and what are involved in building an autonomous learning system that can learn after deployment on the fly. In order to be general, the example is not mainly about dialogue systems but it involves a dialogue system. However, after presenting the SOLA framework, we will briefly discuss a lifelong learning dialogue system that follows the SOLA framework. 


\section{A Motivating Example for SOLA}

The example is about a hotel greeting bot~\citep{chen2018lifelong}. It works in an 
\textbf{open environment} full of unknowns or novel objects, which the classical machine learning cannot handle.  
In general, to make an AI agent such as the greeting bot to thrive in the real open world, like humans, it has to detect novelties and learn them incrementally to make the system more knowledgeable and adaptable over time. It must do so \textit{on its own initiative}\textit{ on the job} (after deployment) rather than relying on human engineers to retrain the system offline periodically. That is, it must learn in the \textit{open world} in a \textit{self-motivated manner} in the context of its \textit{performance task}~(the main task of the agent).

Our hotel greeting bot's \textit{performance task} is to greet hotel guests. When its vision system sees a guest (say, John) it has learned before, it greets him by saying, 

\vspace{+1mm}
``\textit{Hi John, how are you today}?''
\vspace{+1mm}

\noindent
When it sees a new guest, it should detect this guest as new or novel. 
This is a \textbf{\textit{novelty detection}} problem (also known as \textit{out-of-distribution} (OOD) \textit{detection}). Upon discovering the novelty, the new guest, it needs to 
\textbf{\textit{accommodate}} or \textbf{\textit{adapt}} to the novel situation. The bot may say to the new guest

\vspace{+1mm}
``\textit{Hello, welcome to our hotel! What is your name, sir}?''
\vspace{+1mm}

\noindent
If the guest replies ``\textit{David},'' the bot takes some pictures of the guest to \textbf{\textit{gather training data}} and then \textit{\textbf{incrementally or continually learn}} to recognize David. The name ``\textit{David}'' serves as the \textit{class label} of the pictures taken. Like humans, the detected novelty serves as an intrinsic  \textit{self-motivation} for the agent/bot to \textit{learn}. When the bot sees this guest again next time, it may say

\vspace{+1mm}
``\textit{Hi David, how are you today}?''~~~~ (David is no longer novel)
\vspace{+1mm}

In an actual hotel, the situation is, however, much more complex than this. 
For example, how does the system know that the novel object is actually a person, not a dog? If the system can recognize the object as a person, how does it know that he/she is a hotel guest, not a service provider for services such as delivery or security? In order to adapt to the novel object or situation, the system must first \textbf{\textit{characterize}} the novel object, as without it, the agent does not know how to \textbf{\textit{adapt}} or \textit{\textbf{respond}}. In this case, some classification or similarity comparison is needed to decide whether it is a person with luggage. If the object looks like a person but has no luggage, the bot will not respond or learn to recognize the person as it is \textbf{\textit{irrelevant}} to its performance task. If the novel object looks like an animal, it should notify a hotel employee and learn to recognize the object so that it will no longer be novel when it is seen next time. In short, for each \textit{characterization}, there is a corresponding \textit{response} or \textit{adaptation strategy}, which can be NIL (i.e., do nothing). This discussion shows that in order to characterize, the agent must already have a rich volume of world knowledge. Last but not least, there is also \textbf{\textit{risk}} involved when making an incorrect decision. 



The proposed SOLA framework is exactly for dealing with this complex learning environment. Since novelty detection and continual learning are two key issues in SOLA and extensive research has been done about them in the research community, we will review them first.

\section{Novelty Detection}

Novelty detection is also called \textit{out-of-distribution} (OOD) \textit{detection}, \textit{open-world classification} or \textit{open-set recognition}. It is similar to or the same as the earlier \textit{outlier detection} and \textit{anomaly detection} depending on contexts.  Novelty is an agent-specific concept. An object may be novel to one agent based on its partial knowledge of the world but not novel to another agent. We distinguish two types of novelty, \textit{absolute novelty} and \textit{contextual novelty}. 

\vspace{+1mm}
\noindent
\textbf{Absolute novelty.}
Absolute novelty represents something that the agent has never seen before. 
For example, in the 
context of supervised learning, the agent's world knowledge is learned from the training data $\mathcal{D}_{tr}=\{(\boldsymbol{x}_i, y_i)\}_{i=1}^{n}$ with $\boldsymbol{x}_i \in X$ and $y_i \in Y_{tr}$. Let $h(\boldsymbol{x})$ be the latent or internal representation of $\boldsymbol{x}$ in the agent's mind, $h(D_{tr}^i)$ be the latent representation of the training data of class $y_i$, and $k$ ($=|Y_{tr}|$) be the total number of training classes. We use $\mu(h(\boldsymbol{x}), h(D_{tr}^i))$ to denote the novelty score of a test instance $\boldsymbol{x}$ with respect to $h(D_{tr}^i)$. 
The degree of novelty of $\boldsymbol{x}$ with respect to $D_{tr}$, i.e., $\mu(h\boldsymbol{(x)}, h(D_{tr}))$, is defined as the minimum novelty score with regard to every class, 
\begin{equation}
\small
    \mu(h(\boldsymbol{x}), h(D_{tr})) = \min(\mu(h(\boldsymbol{x}), h(D_{tr}^1)), ..., \mu(h(\boldsymbol{x}),h(D_{tr}^k)))
    \label{TCLeq}
\end{equation}

The novelty function $\mu$ can be defined based on specific applications. For example, if the training data of each class follows the Gaussian distribution, one may use the distance from the mean to compute the novelty score. 

\textit{Novel instance}: A test instance $\boldsymbol{x}$ is \textit{novel} or \textit{out-of-distribution} (OOD) if its novelty score $\mu(h\boldsymbol{(x)}, h(D_{tr}))$ is greater than or equal to a threshold value $\gamma$ such that $\boldsymbol{x}$ can be assigned a new class that is not in $Y_{tr}$. 







\textit{Novel class}: A newly created class $y_{new}$ ($y_{new} \notin Y_{tr}$) assigned to some novel instances is called a \textit{novel class} (or \textit{out-of-distribution} (OOD), \textit{unknown} or \textit{unseen class}). The classes in $Y_{tr}$ are called \textit{in-distribution} (IND), \textit{known} or \textit{seen classes}.

\vspace{+1mm}
\noindent
\textbf{Contextual novelty.} 
Based on the prior knowledge of the agent, the probability $P(\boldsymbol{x}|Q)$ of $\boldsymbol{x}$ occurring in a particular context $Q$ is very low, but $\boldsymbol{x}$ has occurred in $Q$, which is \textit{surprising} or \textit{unexpected}. Both $\boldsymbol{x}$  and $Q$ are not absolutely novel as they separately have been seen before. A contextual novelty is also commonly called a \textit{\textbf{surprise}} or \textit{\textbf{unexpected event}}. In human cognition, surprise is an emotional response to an instance which greatly exceeds the expected uncertainty within the context of a task. The definitions of contextual novel instance and class are similar to those for absolute novelty. 

Novelty is not restricted to the perceivable physical world but also includes the agent's internal world, e.g., novel interpretations of world states or internal cognitive states that have no correspondence to any physical world state. Interested readers may also read~\citep{boult2021towards} for a more nuanced and perception-based study of novelty. 



\textbf{Outlier and anomaly}: An outlier is a data point that is far away from the main data clusters, but it may not be unknown. For example, the salary of a company CEO is an outlier with regard to the salary distribution of the company employees, but it is known and thus not novel. Unknown outliers are novel. Anomalies can be considered as outliers or instances that are one off and never repeated. Though technically ``novel'' they may not need to result in a new class.

Note that this book does not deal with various types of data shift such as \textit{covariate shift}, \textit{prior probability shift} and \textit{concept drift} as a large amount of work has been done~\citep{moreno2012unifying}. 

We will not discuss novelty detection further because it has been studied extensively in the literature. Several excellent surveys exist~\citep{pang2021deep,parmar2021open,yang2021generalized}.

\section{Lifelong and Continual Learning}
\label{sec.cl}

Human brains have this extraordinary ability to learn a large number of tasks incrementally with high accuracy. Both the learning process of and the learned knowledge for the tasks have little negative interference of each other. In fact, the learned knowledge earlier can even help the learning of new tasks later. \textit{Continual learning} or \textit{lifelong learning} attempts to make the computer to do the same. The concept of {\em lifelong learning} (LL) was proposed around 1995 
in~\citep{thrun1995lifelong}. Since then, it has been pursued in several directions, e.g., lifelong supervised learning~\citep{chen2018lifelong}, continual learning in deep neural networks~\citep{chen2018lifelong}, lifelong unsupervised learning~\citep{chen2014topic}, lifelong semi-supervised learning~\citep{mitchell2018never}, and lifelong reinforcement learning~\citep{ammar2015autonomous}. LL techniques working in other areas also exist.
\citet{silver2013lifelong} wrote an excellent survey of early LL approaches. A more complete treatment of LL can be found in~\citep{chen2018lifelong}.

The terms \textit{lifelong learning} and \textit{continual learning} have the same meaning and are used interchangeably now, but the past research under the two names has focused on different aspects of the same problem.

The early definition of lifelong learning (LL) in~\cite{chen2018lifelong} as follows, which is based on the early definitions in~\citep{thrun1995lifelong,silver2013lifelong,ruvolo2013ella,mitchell2018never}:

\textbf{Lifelong learning:} At any time point, the learner has learned a sequence of $N$ tasks, $\mathcal{T}_1, \mathcal{T}_2, …, \mathcal{T}_N$. When faced with the $(N+1)^{th}$ task $\mathcal{T}_{N+1}$, the learner can leverage the knowledge learned in the past in the knowledge base (KB) to help learn $\mathcal{T}_{N+1}$. KB maintains the knowledge learned from the previous $N$ tasks. After the completion of learning $\mathcal{T}_{N+1}$, KB is updated with the knowledge gained from learning $\mathcal{T}_{N+1}$. 

We can see the goal of the earlier lifelong learning is to leverage the knowledge learned in the past to learn the new task $\mathcal{T}_{N+1}$ better, i.e., \textit{knowledge transfer}. An implicit assumption of LL is that the tasks learned are very similar~\citep{chen2018lifelong}. The learning setting is almost exclusively the \textit{task continual learning} (TCL) setting (see below), where each task is a separate problem. 

\textbf{Continual learning:} The term \textit{continual learning} (CL) is more commonly used than \textit{lifelong learning} in the deep learning community. Although knowledge transfer is also a goal of CL, the focus of CL has been on solving the \textit{catastrophic forgetting} (CF) problem~\citep{rusu2016progressive,Kirkpatrick2017overcoming,Zenke2017continual,Shin2017continual,serra2018overcoming,lee2019overcoming,chaudhry2020continual,guo2022adaptive,ke2020continual,kim2022continual}. CF refers to the phenomenon that when a neural network learns a sequence of tasks, the learning of each new task is likely to change the weights or parameters learned for previous tasks, which degrades the model accuracy for the previous tasks~\citep{mccloskey1989catastrophic}. 

In the past few years, CF has attracted a great deal of research attention \citep{chen2018lifelong}. There are two main setups in continual learning: \textit{class continual learning} (CCL) (also called \textit{class incremental learning} (CIL)) and \textit{task continual learning} (TCL) (also called \textit{task incremental learning} (TIL) ~\citep{van2019three}. 

\textbf{Class continual learning} (CCL). In CCL, each task consists of one or more classes to be learned together but only one model is learned to classify all classes that have been learned so far. In testing, a test instance from any class may be presented to the model for it to classify with no task information given.
Formally, given a sequence of tasks $\mathcal{T}_1, \mathcal{T}_2, ..., \mathcal{T}_{N}, \mathcal{T}_{N+1}, ...$ and their corresponding datasets $\mathcal{D}_1, \mathcal{D}_2, ..., \mathcal{D}_N, \mathcal{D}_{N+1}, ...$. The dataset of task $\mathcal{T}_k$ is $\mathcal{D}_k=\{(x_k^i, y_k^i)_{i=1}^{n_k}\}$, where $n_k$ is the number of data samples in task $k$, and $x_k^i \in \mathbf{X}$ is an input sample and $y_k^i \in \mathbf{Y}_k$ is its class label. All $\mathbf{Y}_k$'s are disjoint and $\bigcup_{k=1}^T \mathbf{Y}_k = \mathbf{Y}$, where $T$ is the index of the last task $\mathcal{T}_T$ that has been learned.
The goal of CCL is to learn a single prediction function or model $f : \mX \rightarrow \mY$ that can identify the target class $y$ for a given test instance $\vx$. 


\textbf{Task continual learning} (TCL). In TCL, each task is a separate classification problem (e.g., one classifying different breeds of dogs and one classifying different types of birds). TCL builds a set of classification models (one per task) in one neural network. In testing, the system knows which task each test instance belongs to and uses only the model for the task to classify the test instance. Note that classical LL mainly works in this TCL setting and assumes that the tasks are similar to each other to enable knowledge transfer across tasks. Formally, given a sequence of tasks $\mathcal{T}_1, \mathcal{T}_2, ..., \mathcal{T}_{N}, \mathcal{T}_{N+1}, ...$ and their corresponding datasets $\mathcal{D}_1, \mathcal{D}_2, ..., \mathcal{D}_N, \mathcal{D}_{N+1}, ...$.
Each task $\mathcal{T}_k$ has a training dataset
$\mathcal{D}_k=\{((x_k^i, k), y_k^i)_{i=1}^{n_k}\}$, 
where $n_k$ is the number of data samples in task $\mathcal{T}_k \in \mathbf{T} = \{\mathcal{T}_1, \mathcal{T}_2, ..., \mathcal{T}_T\}$, where $\mathcal{T}_T$ is the last task that has been learned, and $x_k^i \in \mathbf{X}$ is an input sample and $y_k^i \in \mathbf{Y}_k \subset \mathbf{Y}$ is its class label.
The goal of TIL is to construct a predictor $f: \mathbf{X} \times \mathbf{T} \rightarrow \mathbf{Y}$ to identify the class label $y \in \mathbf{Y}_k$ for $(x, k)$ (the given test instance $x$ from task $k$).


From now on, we will only use the term \textit{continual learning} (CL) to mean both lifelong learning and continual learning (LL). The goal of CL is to achieve two main objectives, i.e., (1) overcoming CF and (2) performing cross task knowledge transfer. Clearly, not all problems can achieve both. For example, it is not obvious that different tasks or classes can help each other in CCL except some feature sharing. For TCL, if the tasks are entirely different, it is hard to improve the new task learning via knowledge transfer either. For example, one task is to classify whether one has a heart disease or not but another is to classify whether a loan application should be approved or not. 
In these cases, CF is the only problem to solve. 
Recent research has shown that when a mixed sequence of similar and dissimilar tasks are learned in TCL, it is possible to perform selective knowledge transfer among similar tasks~\citep{ke2020continual} and also to overcome CF for dissimilar tasks. Task similarity is detected automatically. 

\begin{figure*}[ht!]
\centering
  \includegraphics[scale=0.80]{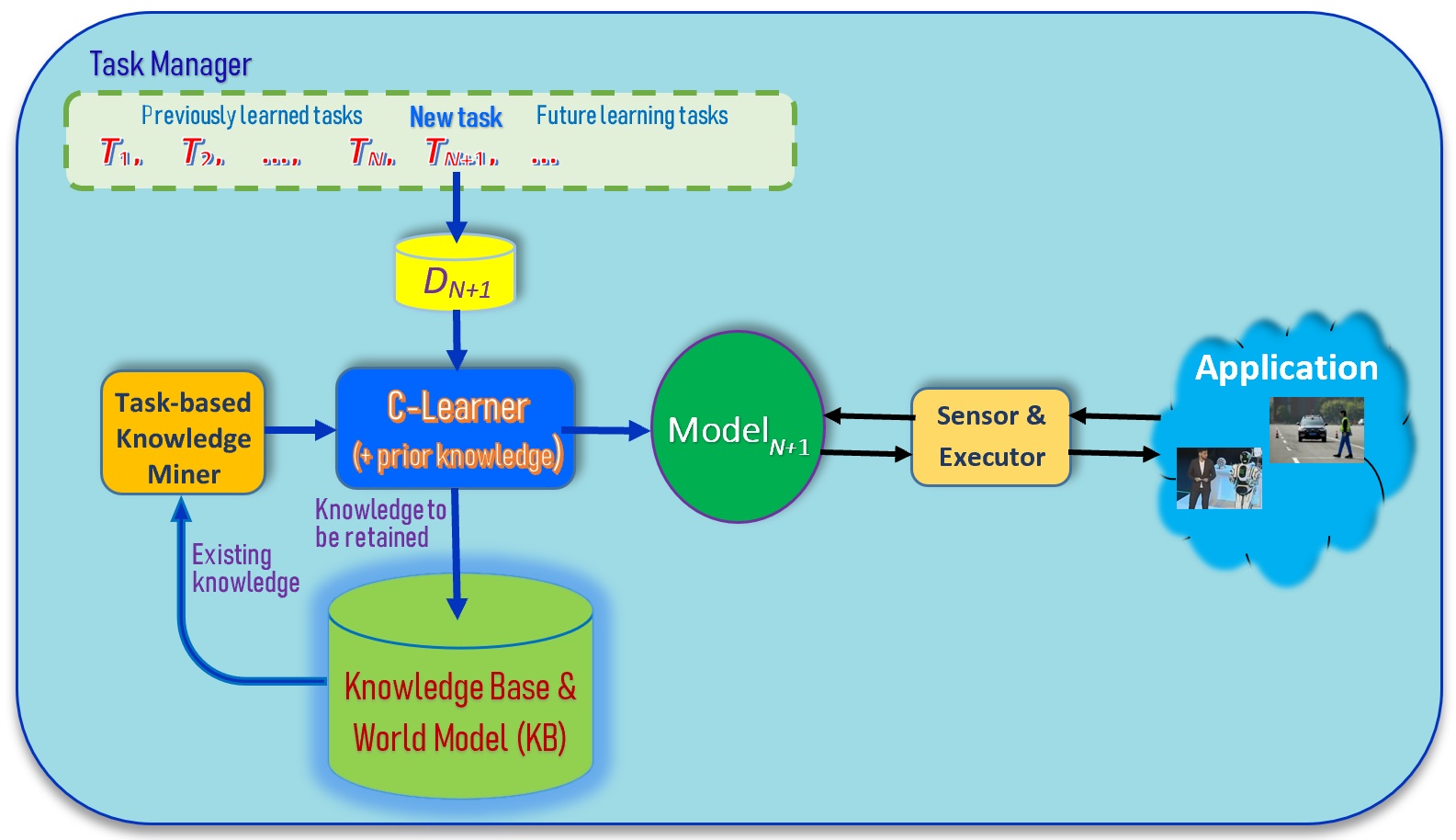}
  \caption{Architecture of a traditional lifelong learning framework.  $\mathcal{T}_1, ..., \mathcal{T}_N$ are the previously learned tasks, $\mathcal{T}_{N+1}$ is the current new task to be learned and $D_{N+1}$ is its training data. The C-Learner (\textit{Continual Learner}) learns by leveraging the relevant prior knowledge identified by the Task-based Knowledge Miner from the Knowledge Base (KB), which contains the retained knowledge in the past. It also deals with the catastrophic forgetting.
}
  \label{fig:traditional_LL}
\end{figure*}

The architecture of CL systems is given in Figure~\ref{fig:traditional_LL}. Apart from those standard components and processes of classical machine learning as explained in Section~\ref{ch2.sec.classic}, this new paradigm has some additions and changes. 

\begin{enumerate}

\item \textbf{C-Learner}: C-Learner means \textit{Continual-Learner}. For CL, it is beneficial for the learner to use prior knowledge in learning the new task. We also call such a learner a {\em knowledge-based learner}, which can leverage the knowledge in the KB to learn the new task. The knowledge relevant to the current is mined by the {\em Task Knowledge Miner} (TKM) (see below). C-Learner will deal with CF in learning a new task. 
\item \textbf{Knowledge Base and World Model (KB)}: It stores the previously learned knowledge and how the world related to the application works.
It may also have a few sub-components:
\begin{enumerate}
\item {\em Past Information Store} (PIS): It stores the information resulted from the past learning, including the resulting models, patterns, or other forms of outcome.
As for what information or knowledge should be retained, it depends on the learning task and the learning algorithm.
For a particular system, the user needs to decide what to retain in order to help future learning. For example, in the \textit{experience-replay} based CL approach, a small number of training examples from each previous task is saved so that they can be used to help deal with CF in learning a new task.   
		\item {\em Knowledge Miner} (MKM). It performs meta-mining of the knowledge in the PIS. We call this {\em meta-mining} because it mines higher-level knowledge from the saved knowledge.
The resulting knowledge is stored in the Meta-Knowledge Store.
Here multiple mining algorithms may be used to produce different types of results. 
		\item {\em Meta-Knowledge Store} (MKS): It stores the knowledge mined or consolidated from PIS (Past Information Store) and also from MKS itself.
Some suitable knowledge representation schemes are needed for each application. 
		\item {\em Knowledge Reasoner} (KR): It makes inference based on the knowledge in MKB and PIS to generate more knowledge.
Most current systems do not have this sub-component.
However, with the advance of CL, this component will become increasingly important.
	
\end{enumerate}
	
Since the current CL research is still in its infancy, none of the existing systems has all these sub-components.

\item {\bf Task-based Knowledge Miner (TKM)}: This module makes use of the raw knowledge or information in the KB to mine or identify knowledge that is appropriate for the current task.
This is needed because in many cases, C-Learner cannot directly use (all) the raw knowledge in the KB. For example, the irrelevant knowledge to the current task needs to be filtered out or blocked. And, C-Learner may only be able to use some more general knowledge mined from the KB~\citep{chen2014topic,ke2020continual}.
	
\item {\bf Model}: This is the learned model, which can be a prediction model or classifier in supervised learning, clusters or topics in unsupervised learning, 
a policy in reinforcement learning, etc. Since the architecture enables continual learning, $Model_{N+1}$ includes all the models from $\mathcal{T}_1$ to $\mathcal{T}_{N+1}$, which may all be in one neural network. In the case of TCL, they may be separate models indexed by their task identifiers. In the case of CCL, $Model_{N+1}$ is just one model that covers all classes of the tasks learned so far. 


\item \textbf{Task Manager (TM)}: It receives and manages the tasks that arrive in the system, and handles the task shift and presents the new learning task to the C-Learner in a lifelong manner.
	
\end{enumerate}

\noindent
{\bf Continual Learning Process}: A typical continual learning process starts with the Task Manager assigning a new task to the C-Learner. C-Learner then works with the help of the past knowledge stored in the KB to produce an output model for the user and also send the information or knowledge that needs to be retained for future use to the KB. Note that dealing with CF is not reflected in the architecture in Figure~\ref{fig:traditional_LL} as it stays in the algorithm of the C-Learner. 

\vspace{+2mm}
\noindent
\textbf{Main Continual Learning Approaches.} Most recent research papers on CL focus on overcoming CF by protecting what the system has learned previously.  There are a large number of existing approaches. They can be roughly grouped into three main categories. 

\textit{Regularization-based approaches}:
The main idea of these approaches is to compute the importance of each parameter or gradient to the previously learned tasks, and then add a regularization in the loss function to penalize changes to those important parameters to prevent CF on previous tasks~\citep{jung2016less,Camoriano2017incremental,lee2019overcoming,schwarz2018progress,ahn2019uncertainty,zhu2021prototype,dhar2019learning}. 
For example,
EWC~\citep{Kirkpatrick2017overcoming}, one of the most popular algorithms, uses the Fisher information matrix to represent the importance of parameters. This technique works to some extent, but is very weak at dealing with CF. SI~\citep{Zenke2017continual} was proposed to extend EWC so that it has less complexity in computing the penalty. Several approaches
\citep{li2017learning,buzzega2020dark} also use knowledge distillation \citep{hinton2015distilling} to preserve the previous knowledge. 

\textit{Replay-based approaches}:
The idea of these approaches is to use a small memory buffer to save a small amount of data from previous tasks and replay or use them to jointly train the new task together with the new task data to prevent forgetting of the knowledge learned from previous tasks~\citep{rusu2016progressive,Lopez2017gradient,rebuffi2017icarl,chaudhry2019efficient,rolnick2019experience, rajasegaran2019adaptive,liu2021adaptive,cha2021co2l,buzzega2020dark}. This approach is also called \textit{experience replay} or the \textit{memory-based} approach. Example systems include GEM \citep{Lopez2017gradient} and A-GEM \citep{chaudhry2019efficient}. Instead of saving some previous data, some approaches learn a data generator for previous tasks~\citep{Gepperth2016bio,Kamra2017deep,Shin2017continual,Seff2017continual,Kemker2018fearnet,hu2019overcoming,ostapenko2019learning}. In learning a new task, the generator generates pseudo-samples of previous tasks and  uses them instead of real samples to jointly train the new task. 

\textit{Parameter isolation-based} (or \textit{architectural approaches}:
These approaches are mainly used in task-incremental learning (TIL). Its main idea is to learn a sub-network for each task, and tasks may share some parameters and neurons~\citep{serra2018overcoming,ke2020continual}. 
HAT~\citep{serra2018overcoming} and SupSup~\citep{wortsman2020supermasks} are two representative systems.
HAT learns neurons (not parameters) that are important for each task and ``hard masks'' them via the task embeddings based on gates. SupSup uses a different approach but it also learns and fixes a sub-network for each task. Besides them, many other systems also take similar approaches, e.g., Progressive Networks~\citep{rusu2016progressive}, PackNet~\citep{mallya2018packnet}, HyperNet~\citep{von2020continual}, and BNS~\citep{qin2021bns}.

Several recent works have also attempted to deal with knowledge transfer and catastrophic forgetting (CF) together. For example B-CL~\citep{ke2021adapting} and CTR~\citep{ke2021achieving} train a shared adapter for all tasks in a pre-trained language model. They prevent CF via task masks~\citep{serra2018overcoming} (parameter isolation) and achieve knowledge transfer via capsule networks~\citep{sabour2017dynamic}. CLASSIC~\citep{ke2021Classic} achieves knowledge transfer via contrastive learning. CAT~\citep{ke2020continual} learns a sequence of similar and dissimilar tasks and it also deals with both CF and knowledge transfer. 


\vspace{+2mm}
\noindent
\textbf{Limitations.} One key limitation of the existing CL paradigm is that the tasks and their training data are given by the user. This means that the system is not autonomous and cannot learn by itself. In order to do that, we extend the CL architecture in Figure~\ref{fig:traditional_LL} to enable open-world on-the-job learning to achieve full SOLA.  

\section{The SOLA Framework} \label{sec.SOLA}

\begin{figure}[t!]
\centering
  \includegraphics[scale=0.80]{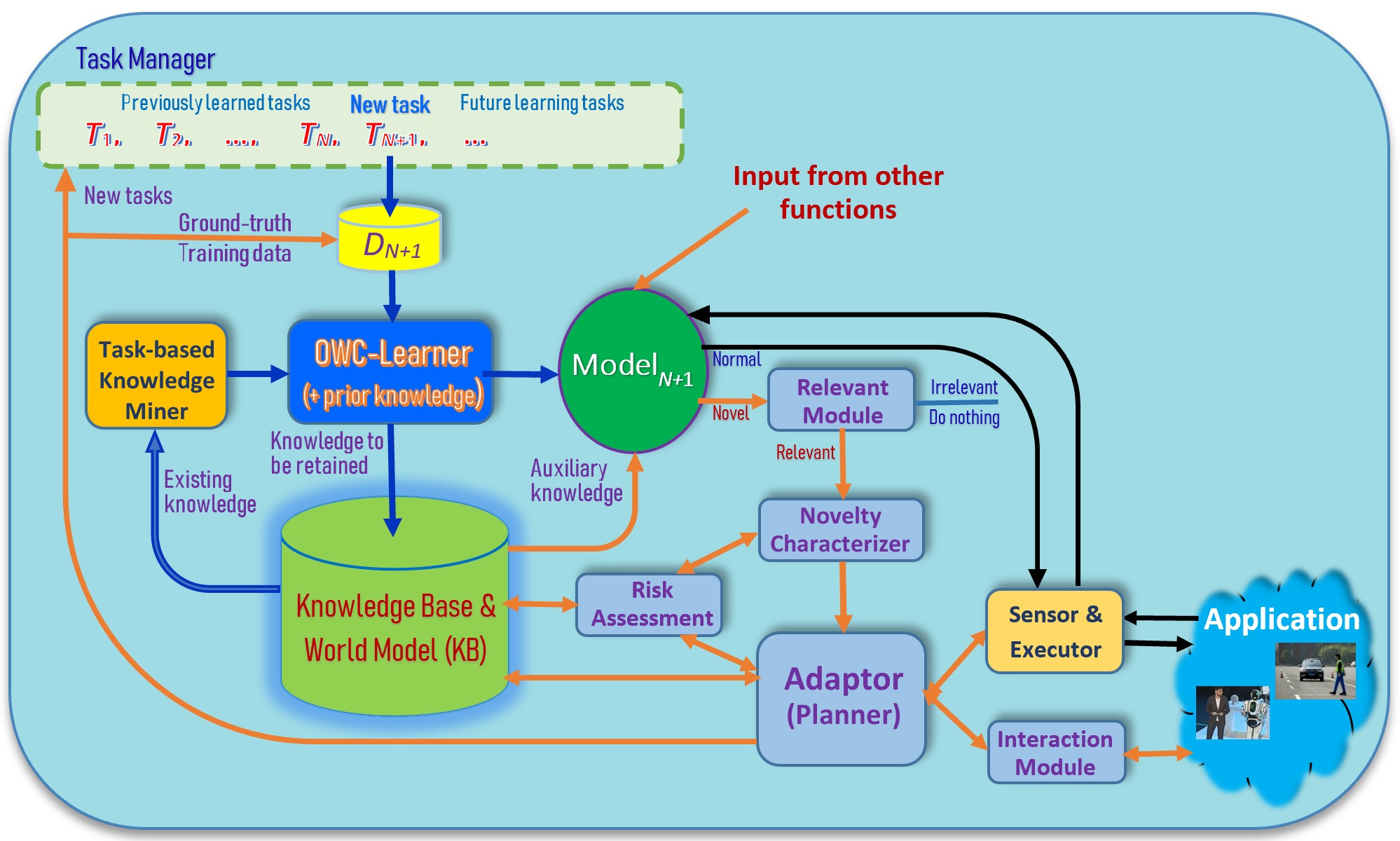}
  \caption{Architecture of the primary task performer or any supporting function. 
  OWC-Learner means \textit{Open-World Continual Learner}. 
  }
  \label{fig:arch-SOLA}
\end{figure}

The SOLA architecture is given in Figure~\ref{fig:arch-SOLA}, which adds the orange-colored links and associated components to the continual learning architecture in Figure~\ref{fig:traditional_LL}. 
These newly added links and components enable the system to learn by itself to achieve autonomy, which is what SOLA aims to achieve. It is called \textbf{learning after deployment} or \textbf{learning on the job} during application or after model deployment.

\vspace{2mm}
\noindent
\textbf{Learning after deployment} refers to learning after the model has been deployed in an application or during model application~\citep{liu2021lifelong} on the fly. 
The basic idea is that during application, if the system/agent encounters anything that is \textit{out-of-distribution} (OOD) or novel, it needs to \textbf{detect the novelty}. Based on the novelty, the system \textbf{creates a new task} to learn and also acquires the \textbf{ground-truth training data} to learn the task on the initiation of the system itself through interactions with humans and the environment. The system then \textbf{learns the new task incrementally} or continually. The whole process is carried out on the fly during application. 

\subsection{Components of SOLA}

SOLA is proposed as a framework for autonomous AI agents. An AI agent consists of a pair of key modules $(P, S)$, where $P$ is the \textit{primary task-performer} that performs its performance task (e.g., the dialogue system of the greeting bot) and $S$ is \textit{a set of supporting or peripheral functions} (e.g., the vision system and the speech system of the bot) that supports the primary task-performer. The primary task-performer $P$ or each supporting function $S_i \in S$ consist of eight core sub-systems 
($L, M, K, R, C, A, S, I$).  Figure~\ref{fig:arch-SOLA} shows the relationships and functions of the sub-systems.  We do not distinguish $P$ and $S_i$ in terms of techniques or subsystems as we believe they have no fundamental difference. 
\vspace{-1mm}
\begin{itemize}
\item \textbf{$L$} is an \textbf{OWC-Learner} (\textit{Open-World Continual Learner}) that builds models to not only classify the input into known classes but also detect novel objects that have not been seen in training. For example, for the greeting bot, $L$ of the primary task performer $P$ is a continual learning dialogue system similar to that in Section~\ref{sec.cml}. For the supporting vision system, $L$ is a continual learner that can learn to recognize guests and detect novel or unknown objects. 
Compared to C-Learner in continual learning in Figure~\ref{fig:traditional_LL}, OWC-Learner in SOLA not only can learn continually like C-Learner but also produce models that can detect novel instances in testing or in application deployment. 

\item \textbf{$K$} is the\textbf{ Knowledge Base \& World Model} (KB) that is important for the performance task, supporting functions or the OWC-Learner. KB is also in Figure~\ref{fig:traditional_LL} but plays more roles in SOLA. Apart from keeping the learned or prior knowledge of the domain and the world model, if needed, reasoning capability may also be  provided to help the other modules of the system (see the orange-colored links). Some knowledge from the application observed by the Adapter (see below) may be added to the KB, which can provide some knowledge to the Model for its decision making. World model refers to the representation of the task environment and the commonsense knowledge about the objects and their relationships within. 
\item \textbf{$M$} is the \textbf{Model} learned by $L$. $M$ takes the input or perception signals from the application environment to make a decision to perform actions in the application. $M$ may also use some input or knowledge from other supporting functions.  
\item \textbf{$R$} is the \textbf{Relevance Module} or focusing mechanism that decides whether the detected novelty is relevant to the current task or not. If it's relevant, the agent should respond to the novelty (discussed below); otherwise ignore it. For example, in the greeting bot application, when the bot hears something from people who are chatting with each other, whether understandable or not, it will ignore them as they are irrelevant to its performance task.      
\item \textbf{$C$} is the \textbf{Novelty Characterizer} that characterizes the detected novelty based on the knowledge in the KB so that the adaptor (below) can formulate a course of actions to respond or adapt to the novelty. For the characterizer $C$ of $P$ of the greeting bot, as $P$ is a dialogue system, when it cannot understand the utterance of a hotel guest (a novelty), it should decide what it can and cannot understand (see Section~\ref{sec.cml}) and ask the guest based on its partial understanding (see below). {\color{black}In the case of the supporting vision system, when a novel object it detected, the charaterizer may decide what the object looks like and its physical attributes. For example, the novel object may look like a dog based on the greeting bot's KB (see Section~\ref{sec.char} for more discussions)}.
\item \textbf{$A$} is the \textbf{Adaptor} that adapts to or accommodates the novelty based on the characterization result. It is a \textit{\textbf{planner}} that produces a plan of actions for the executor $E$ or the interactive module $I$ to perform. Given the characterization (e.g., partial understanding) above, $A$ may adapt by asking the guest to clarify (see Section~\ref{sec.cml}) and then learn to understand the utterance. In the case of the vision system, if the characterizer believes that the novel object looks like a dog, the adaptor may decide to report to a hotel employee and then learns the new object by taking some pictures as the training data and asking the hotel employee for the name of the object as the class label. In the latter two cases, $A$ needs to invoke $I$ to interact with the human and $L$ to learn the novelty so that it will not be novel in the future. That is, $A$ is also responsible for creating new tasks (e.g., learning to recognize new objects by the greeting bot) on the fly and proceeds to acquire ground truth training data with the help of $I$ (discussed below) to be learned by $L$. This adaptation process often involves reasoning.
\item $S$ is the \textbf{Risk Assessment} module. Novelty implies uncertainty in adapting to the novel situation. In making each response decision, risk needs to be assessed (see Section~\ref{sec.risk} for more discussions).
\item \textbf{$I$} is the \textbf{Interactive Module} for the agent to communicate with humans or other agents, e.g., to acquire ground-truth training data or to get instructions when the agent does not know what to do in a unfamiliar situation. It may use the natural language (for interaction with humans) or an agent language (for interaction with other agents).

\vspace{-1mm}
\end{itemize}


Several remarks are in order. First, not all agents need all these sub-systems and some sub-systems may also be shared. For example, the primary task performer $P$ in the greeting bot application is a dialogue system. Its interaction module $I$ can use the same dialogue system. In some cases, the Model may also be able to determine the relevance of a novel object to the application and even characterize the novelty because characterization in many cases is about classification and similarity comparison. Second, as we will see, every sub-system can and should have its own local learning capability. Third, the interaction module $I$ and the adapter $A$ will create new tasks to learn and gather ground truth training data for learning. Fourth, most links in Figure~\ref{fig:arch-SOLA} are bidirectional, which means that the sub-systems need to interact with each other in order to perform their tasks. The interactions may involve requesting for information, passing information, and/or going back and forth with hypothesis generation and hypothesis evaluation to make more informed decisions. 

Since the primary task performer $P$ and each supporting sub-system $S_i$ has the same components or sub-systems, we will discuss them in general rather than distinguishing them. Below, we first discuss the SOLA learner $L$.

\subsection{Open World Continual Learning}\label{sec.open-world-continual-learning}

As discussed earlier, the classical ML makes the \textit{i.i.d} assumption, which is often violated in practice. Here we first define some concepts and then present the paradigm of \textit{open-world continual learning} (OWC-Learning).

Let the training data that have been seen so far from previous tasks be $\mathcal{D}_{tr}=\{(\boldsymbol{x}_i, y_i)\}_{i=1}^{n}$ with $\boldsymbol{x}_i \in \mathbf{X}$ and $y_i \in \mathbf{Y}_{tr}$. Let the set of class labels that may appear in testing or application be $\mathbf{Y}_{tst}$. Classical ML makes the closed-world assumption. 

\textbf{Closed-world assumption}: There are no new or novel instances or classes that may appear in testing or application, i.e., $\mathbf{Y}_{tst} \in \mathbf{Y}_{tr}$. In other words, every class seen in testing or application must have been seen in training.  

\textbf{Open world:} There are test classes that have not been seen in training, i.e., $\mathbf{Y}_{tst} - \mathbf{Y}_{tr} \neq \emptyset$.

\textbf{Definition (closed-world learning)}: It refers to the learning paradigm that makes the closed-world assumption. 




\textbf{Definition (open world learning (OWL))}: It refers to the learning paradigm that performs the following functions: (1) classify test instances belonging to training classes to their respective classes and detect 
novel or out-of-distribution instances, and 
(2) learn the novel classes labeled by humans in the identified novel instances to update the model using the labeled data. The model updating is initiated by human engineers and involves re-training or incremental learning. 


\textbf{Definition (open-world continual learning (OWC-learning)):} {\color{black} OWC-learning is the learning paradigm that performs open-world learning but the learning process is initiated by the agent itself after deployment with no involvement of human engineers. The new task creation and ground-truth training data acquisition are done by the agent via its interaction with the user and the environment. The learning of the new task is incremental, i.e., no re-training of previous tasks/classes. The process is lifelong or continuous, which makes the agent more knowledgeable over time.} 


\subsubsection{Steps in OWC-Learning} The main process in OWC-Learning involves the following three steps, which can be regarded as part of the novelty adaptation or accommodation strategy (see Section~\ref{sec.char}). 

\noindent
\textbf{Step 1} - \textit{Novelty detection}. This step involves detecting data instances whose classes do not belong to $\mathbf{Y}_{tr}$.  As mentioned earlier, a fair amount of research has been done on this under open-set classification or out-of-distribution (OOD) detection~\citep{pang2021deep}.  

\noindent
\textbf{Step 2} - \textit{Acquiring class labels and creating a new learning task on the fly}: This step first clusters the detected novel instances. Each cluster represents a new class. Clustering may be done automatically or through interaction with humans using the interaction module $I$. Interacting with human users should produce more accurate clusters and also obtain meaningful class labels. If the detected data is insufficient to learn an accurate model to recognize the new classes, additional ground-truth data may be collected via interaction with human users (and/or passively by downloading data from web, like searching and scrapping images of objects of a given class). A new learning task is created. 

In the case of our hotel greeting bot, since the bot detects a single new guest (automatically), no clustering is needed. It then asks the guest for his/her name as the class label. It also takes more pictures as the training data. With the labeled ground-truth data, a new learning task is created to incrementally learn to recognize the new guest on the fly.  

The learning agent may also interact with the environment to obtain training data. In this case, the agent must have an \textbf{\textit{internal evaluation system}} that can assign rewards to different states of the world, e.g., for reinforcement learning. 

\noindent
\textbf{Step 3} - \textit{Incrementally learn the new task.} After ground-truth training data has been acquired, the OWC-Learner $L$ incrementally learns the new task. This is \textit{continual learning}. We will not discuss it further as there are already numerous existing techniques (see Section~\ref{sec.cl}). Many can leverage existing knowledge to learn the new task better~\citep{chen2018lifelong}.  

\subsubsection{Acquiring Training Data Automatically} 
Existing approaches to obtaining the training data is through manual labeling or writing (in the case of dialogue data), which is both costly and time-consuming. In the case of chatbots, as they typically work in multi-user environments, we can exploit such environments to acquire the ground truth training data interactively during actual online conversations. This process is both automatic and free. We have discussed this issue in Section~\ref{ch1.sec4} and we briefly reproduce it here for completeness of this section. 


\textbf{1. Extracting data or information directly from user utterances} 
(or dialogue history), which can be real-world facts, user preferences, etc. 
Additional knowledge/data may be inferred from the acquired and existing KB knowledge. 


\textbf{2. Asking the current user} when the agent (1) doesn't understand a user utterance, or (2) cannot answer a user query, which forms a new learning task. To obtain the ground truth data, for (1), the agent may ask the current user for clarification or rephrasing. For (2), it may ask the user for some supporting facts and then infer the query answer~\citep{mazumder2019lifelong,mazumder2020continuous}. In this process, it obtains command-action pairs, question-answer pairs, etc.


\textbf{3. Asking other users} to obtain the answers when the chatbot could not answer a user query. For example, if a user asks  ``\textit{What is the capital city of the US?}" and the chatbot is unable to answer or infer now, it can try to find a good opportunity to ask another user later ``\textit{Hey, do you happen to know what the capital city of the US is?}" If the user gives the answer ``\textit{it's Washington DC}," the agent acquires the ground truth (a piece of new knowledge) which can be used in its future conversations or as a piece of training data.  

\textbf{4. Observing user demonstrations if supported}. See Section~\ref{ch1.sec4} and also~\citep{mazumder2020application}. 

\vspace{+1mm}
\noindent
Beyond these, the agent may also extract ground-truth data from online documents or online knowledge bases.






\subsection{Relevance of Novelty}
As an AI agent has a performance task, it should focus on novelties that are critical to the performance task. For example, a self-driving car should focus on novel objects or events that are or may potentially appear on the road in front of the car. It should not pay attention to novel objects in the shops along the street (off the road) as they do not affect driving. This relevance check involves gathering information about the novel object to make a classification decision. As this is a normal classification task, it is not discussed further. 

\subsection{Novelty Characterization and Adaptation} \label{sec.char}
In a real-life application, classification may not be the primary task of the agent. 
For example, in a self-driving car, object classification or recognition supports its primary performance task of driving. 
To drive safely, the car has to take some actions to adapt or respond to any novel/new objects, e.g., slowing down and avoiding the objects. In order to know what actions to take to adapt, the agent must characterize the new object. The \textbf{characterization} of a novel object is a partial description of the object based on the agent's existing knowledge about the world. {\color{black}According to the characterization, appropriate actions are formulated to \textbf{adapt} or respond to the novel object. The process may also involve learning.} 











\textit{Novelty characterization} and \textit{adaptation} (or {response}) form a pair $(c, r)$, where $c$ is the \textit{characterization of the novelty} and $r$ is the adaptation \textit{response} to the novelty, which is a plan of dynamically formulated actions based on the characterization of the novelty. The two activities go hand-in-hand. Without an adaptation strategy for a characterization, the characterization has little use. If the system cannot characterize a novelty, it takes a low risk-assessed \textit{default response}. 
In our greeting bot example, when it can characterize a novelty as a new guest, its response is to say "\textit{Hello, welcome to our hotel! What is your name, sir}?" If the bot has difficulty with characterization, it can take a \textit{default action}, e.g., `{do nothing}.'  The set of responses are specific to the application. For a self-driving car, the default response to a novel object is to slow down or stop the car so that it will not hit the object.
In some situations, the agent must take an action under low confidence circumstances, e.g., the agents engage in reinforcement learning, trying actions and assessing outcomes.


Characterization can be done at different levels of detail, which may result in more or less precise responses. Based on an ontology and object attributes related to the performance task in the domain, characterization can be described based on the \textit{type of the object} and the \textit{attribute of the object}. For example, in the greeting bot application, it is useful to determine whether the novel object is a human or an animal because the responses to them are different. 
For self-driving cars, when sensing a novel object on the road, it should focus on those aspects that are important to driving, i.e., whether it is a still or a moving object. If it is a moving object, the car should determine its direction and speed of moving. Thus, classification of movement is needed in this case to characterize the novelty, which, in turn, facilitates determination of the agent's responding action(s). For instance, if the novel object is a mobile object, the car may wait for the object to leave the road before driving. 

Another characterization strategy is to \textit{compare the similarity} between the novel object and the existing known objects. For example, if it is believed that the novel object looks like a dog (assuming the agent can recognize a dog), the agent may react like when it sees a dog on the road. 


The above discussion implies that in order to effectively characterize a novelty, the agent must already have a great deal of world knowledge that it can use to describe the novelty. Additionally, the characterization and response process is often interactive in the sense that the agent may choose a course of actions based on the initial characterization. After some actions are taken, it may get some feedback from the environment. Based on the feedback and the agent's additional observations, the course of actions may change. 

\textbf{Learning to respond.} In some situations, the system may not know how to respond to a novel object or situation. It may try one or more of the following ways. 

(1) \textit{Asking a human user}. In the case of the self-driving car, when it does not know what to do, it may ask the passenger using the interactive module $I$ in natural language and then follow the instruction from the passenger and also learn it for future use. For example, if the car sees a black patch on the road that it has never seen before, it can ask ``\textit{what is that black thing in front?}'' The passenger may answer ``\textit{that is tar}.''  If there is no ready response, e.g., no prior information on tar, the system may progress with a further inquiry, asking the passenger ``\textit{what should I do?}''  

(2) \textit{Imitation learning}. On seeing a novel object, if the car in front drives through it with no issue, the car may choose the same course of action as well and also learn it for future use if the car drives through without any problem. 

(3) \textit{Reinforcement learning}. By interacting with the environment through trial and error exploration, the agent learns a good response policy. This is extremely challenging as any action taken has consequences and cannot be reversed. For this to work, the agent must have an internal evaluation system that is able to assign rewards to (possible) states.  

If multiple novelties are detected at the same time, it is more difficult to respond as the agent must reason over the characteristics of all novel objects to dynamically formulate an overall plan of actions that prioritize the responses.


\subsection{Risk Assessment and Learning}
\label{sec.risk}
There is risk in achieving performance goals of an agent when making an incorrect decision. For example, classifying a known guest as unknown or an unknown guest as known may negatively affect guest impressions resulting in negative reviews. For a self-driving car, misidentifications can result in wrong responses, which could be a matter of life and death. Thus, risk assessment must be done in making each decision. Risk assessment can also be learned from experiences or mistakes.  In the example of a car passing over tar, after the experience of passing over shiny black surfaces safely many times, if the car slips in one instance, the car agent must assess the risk of continuing the prior procedure.  Given the danger, a car may weight the risk excessively, slowing down on new encounters of shiny black surfaces. 

\textbf{Safety net}: Another aspect of risk is safety net. If a wrong decision is made, it should not cause a catastrophe. A wrong decision may be due to two reasons. The first is the model inaccuracy. For example, the perception system of a self-driving car may classify a pedestrian wearing a black dress as a patch of tar on the surface of the road, and run over the person. The second is wrong knowledge acquired from human users during human-agent interaction. For example, 
if a self-driving car sees an unknown object (e.g., a big stone) in the middle of the road and asks the passenger what it is and what to do, the passenger may say it is a stone and it is safe to drive through it. In this case, the car should not go ahead and hit the stone.

\section{A Dialogue System based on SOLA} \label{sec.cml}
We now briefly describe a dialogue system (called CML) in~\citep{mazumder2020application} that uses the SOLA framework and learns continually by itself on the fly during conversation with users to become more and more powerful. Details will appear in Section~\ref{ch4.sec2}. Another two related systems~\citep{mazumder2019lifelong,mazumder2020continuous} can be found in Section~\ref{ch3.sec4}, which are more sophisticated than CML. 

CML is a natural language interface system like Amazon Alexa and Apple Siri. Its \textbf{performance task} is to take a user command in natural language (NL) and perform the user requested API action in the underlying application, e.g., ``\textit{switch off the light in the bedroom}.'' Since CML is a text-based research system, it does not have any other support functions or sub-systems. The key issue of CML is to understand paraphrased NL commands from the user in order to map a user command to a system’s API call. The \textbf{Model} of CML consists of two parts. The first part is a store $S$ of \textit{seed commands} (SCs). Each SC in $S$ is similar to an actual user command except that the arguments used by the associated API function is replaced by variables, e.g., ``\textit{switch off the light in the $X$},'' where $X$ is the variable representing the argument of the API function for this command. A SC is either prepared by the API developer before the system is deployed or continually learned from users on the fly during conversation. The second part is a command matching system that matches a user command to a saved SC in $S$. If no match can be found, it means an unexpected or \textit{novel command} from the user is detected, i.e., a \textit{novelty}, which equates to the system's failure in understanding or grounding a user command. The novelty detection task is performed by the \textbf{Model}. CML assumes that every novelty is \textbf{relevant} to the performance task, but if needed, a \textbf{Relevance Module} can be built as a classifier to classify whether a user command is relevant or not to the application.
\textbf{Novelty characterization} of CML is also performed by the \textbf{Model}. In the matching process, CML tries to identify the part of the user command that the system does not understand and how similar it is to some known SCs in $S$. Based on the characterization, the \textbf{Adaptor} dynamically formulates an interaction strategy based on the context (e.g., dialogue history, command from the user, information acquired from user and outstanding information needed to complete a task, etc.) and executes the strategy to carry out multi-turn dialogues for knowledge acquisition (i.e, obtaining the ground truth data from the user). Specifically, 
in CML, it decides \textit{when to ask} and \textit{what to ask}  the user, to achieve two goals: (1) understand the user command so that it can perform the correct API action for the user, and (2) create a new task and use the correct API action as the ground-truth for the user command for learning by \textbf{OWC-Learner}. CML uses a rule-based system for adaptation. Note that a finite state machine is used in another system~\citep{mazumder2020continuous}.   
Based on the user's command and the ground-truth API action, \textbf{OWC-Learner} learns this command incrementally by creating a new SC and adding it to the SC store $S$ so that the system will be able to match this and similar commands in the future. 
In the adaptation or accommodation process, \textbf{risk} is also considered (see below). There is no separate \textbf{Interaction Module} because the main task performer is a dialogue system, which is an interactive system itself and can serve the role of the Interaction Module. 

    

Consider the following example user command ``\textit{turn off the light in the kitchen}'' that the \textbf{Model} cannot match or ground to an existing SC (i.e., it does not understand). A \textbf{novelty} is reported. The \textbf{Model} also \textbf{characterizes} the novelty by deciding which part of the command it can match or understand, which part it has difficulty with, and what known SCs are similar to the user command. Based on the characterization result, the \textbf{Adapter} formulates actions to be taken. In this case, it provides the user a list of \textit{top-k} predicted actions (see below) described in natural language and asks the user to select the most appropriate action from the given list. 
\begin{quote}
    \vspace{-2mm}
    \small
    \rule{0.9\linewidth}{0.4pt}\\
    \textbf{Bot}: Sorry, I didn't get you. Do you mean to:\\
    \vspace{-2mm}\\
    \textbf{option-1.}~~ switch off the light in the kitchen, or\\ \textbf{option-2.}~~ switch on the light in the kitchen, or \\ 
    \textbf{option-3.}~~ change the color of the light?\\
    \rule{0.9\linewidth}{0.4pt} 
    \vspace{-1mm}
\end{quote}
The user selects the desired action (option-1). The action API [say, \texttt{SwitchOffLight}(\textit{arg}:place)] corresponding to the selected action (option-1) is retained as the ground truth action for the user-issued command. The \textbf{Adaptor} invokes the \textbf{Executor} to perform the API action for the user. In subsequent turns of the dialogue, the system can also ask the user questions to acquire ground truth values associated with arguments of the selected action, as defined in the API. Based on the acquired information, OPC-Learner of CML \textbf{incrementally learns} to map the original command ``\textit{turn off the light in the kitchen}'' to the API action,  \texttt{SwitchOffLight}(\textit{arg}:place), by creating a new SC (``\textit{turn off the light in the X}'') and adding it to the SC store, which ensures that in the future the system will not have problem understanding the related commands.  


\textbf{Risk} is considered in CML in two ways. First, it does not ask the user too many questions in order not to annoy the user. Second, when the characterization is not confident, the \textbf{Adapter} simply asks the user to say his/her command again in an alternative way, i.e., ''\textit{Can you say it again in another way}?'' (the new way may be easier for the system to understand) rather than providing a list of random options for the user to choose from. If the options have nothing to do with the user command, the user may lose confidence in the system.

\vspace{+1mm}

\section{Comparison with Related Work} 
\label{sec.related-work}
This section summarizes the difference between the SOLA framework and some existing machine learning frameworks or paradigms. There are three machine learning paradigms that are closely related to SOLA, i.e., \textit{novelty detection}, \textit{open-world learning}, and \textit{continual learning}. We compare SOLA with these three topics below. 

\subsubsection{Novelty Detection}
Novelty detection is also known as \textit{anomaly detection},  \textit{out-of-distribution (OOD) detection}, \textit{open set detection}, or \textit{open classification}. Extensive research has been done on novelty detection~\citep{xu2019open,fei2016breaking}. Two recent surveys of the topic can be found in~\citep{yang2021generalized,pang2021deep}. Novelty detection is only the first step in SOLA. Novelty detection does not involve lifelong or continual learning, which is the backbone of SOLA, and it also does not involve novelty characterization and adaptation or the associated risk assessment. 

\subsubsection{Open-World Learning}
Some researchers have also studied learning the novel objects after they are detected and manually labeled~\citep{bendale2015towards,fei2016learning,xu2019open}, which are called \textit{open-world learning}. A survey of the topic is given in~\citep{parmar2021open}.
A position paper~\citep{langley2020open} presented some blue sky ideas about open-world learning, but it does not have sufficient details or an implemented system. SOLA differs from open-world learning in many ways. The key difference that in open-world learning, the tasks and their trained data are given by the user or engineers. SOLA stresses self-initiation in learning, which means that all the learning activities from start to end are \textit{self-motivated} and \textit{self-initiated} by the agent itself. 
The process involves no human engineers, which is important for AI autonomy. Furthermore, the research in open-world learning so far has not involved continual learning of a sequence of tasks and thus has not dealt with the issue of catastrophic forgetting and knowledge transfer. 

Due to self-initiation, SOLA enables learning after the model deployment like human learning on the job or while working, which has barely been attempted before. In existing learning paradigms, after a model has been deployed, there is no more learning until the model is updated or retrained on the initiation of human engineers. 


\subsubsection{Continual Learning}
Continual learning (CL) aims to learn a sequence of tasks incrementally. However, existing research assumes that the tasks and their training data are given by users or engineers. SOLA differs from CL in two main ways. 

\textbf{(1).} SOLA includes modules to characterize and adapt to novel situations so that the agent can work in the open world environment without stopping. These are not included in continual learning. 

\textbf{(2).} SOLA makes learning autonomous and self-initiated. This involves online interactions of the learning agent with human users, other AI agents, and the environment. The purpose is to acquire ground-truth training data on the fly by itself. This is very similar to what we humans do when we encounter something novel or new and ask others interactively and learn it. It is very different from collecting a large amount of unlabeled data and asking human annotators to label the data. Incidentally, SOLA also differs from active learning \citep{settles2009active,ren2021survey} as active learning only focuses on acquiring labels from users for selected unlabeled examples in a given dataset. 




Finally, note that although SOLA focuses on self-initiated continual learning, it does not mean that the learning system cannot learn a task given by humans or other AI agents. {\color{black}Additionally, SOLA also allows learning from other resources, e.g., the Web, to gain new knowledge, like a human reading a book. We will describe the system NELL~\citep{mitchell2018never} that continuously learns new knowledge by extracting facts from Web documents in Section~\ref{ch3.sec5}.}


\section{Summary}
\label{ch2.summary}

This chapter presented a new framework called \textit{self-initiated open world continual learning and adaptation} (SOLA) that should  be followed by AI agents (including chatbots or dialogue systems) that want to continually learn after model deployment or learn on the job. We believe that this capability is necessary for the next generation machine learning or AI agents and any form of intelligence in general. The core of this framework is self-motivation and self-initiation. That is, the AI agent must learn autonomously and continually in the open world on its own initiative after deployment (or post-deployment), detecting novelties, adapting to the novelties and the ever-changing world, and learning more and more to become more and more powerful over time. An example SOLA based dialogue  system (called CML) is briefly described (see Section~\ref{ch4.sec2} for more details of the system). Almost all existing dialogue systems related to lifelong learning are only partial continual learning dialogue systems. They do not include all functions in the proposed framework. For example, some only perform continual learning with given tasks and training data, some only detect novelties, and some only continually collect additional training data during chatting.



%% file: ch03.tex
\graphicspath{{./figures/Ch03}}     

\chapter{Continuous Factual Knowledge Learning in Dialogues}
\label{ch3}

\section{Opportunities for Knowledge Learning in Dialogues}
\label{ch3.sec1}

\section{Extracting Facts from Dialogue Context}
\label{ch3.sec2}

\section{Lexical Knowledge Acquisition in Dialogues}
\label{ch3.sec3}

\section{Interactive Factual Knowledge Learning and Inference}
\label{ch3.sec4}

\section{Learning new knowledge from external sources}
\label{ch3.sec5}

\section{Summary}
\label{ch3.sec6}

%% file: ch04.tex
\graphicspath{{./figures/Ch04}}     

\chapter{Continuous and Interactive Language Learning and Grounding}
\label{ch4}

\section{Modes of language learning human-chatbot interactions}
\label{ch4.sec1}

\section{Learning language games through interactions}
\label{ch4.sec2}

\section{Dialogue-driven Learning of Self-adaptive NLIs}
\label{ch4.sec3}

\section{Interactive semantic parsing and learning from feedback}
\label{ch4.sec4}

\section{Summary}
\label{ch4.sec5}

%% file: ch05.tex
\graphicspath{{./figures/Ch05}}     

\chapter{Continual Learning in Chit-chat Systems}
\label{ch5}

\section{Predicting User Satisfaction in Open-domain Conversation}
\label{ch5.sec1}

\section{Learning by Extracting New Examples from Conversation}
\label{ch5.sec2}

\section{Dialogue Learning via Role-Playing Games}
\label{ch5.sec3}

\section{Summary}
\label{ch5.sec4}

%% file: ch06.tex
\graphicspath{{./figures/Ch06}}     

\setcounter{algorithm}{0}

\chapter{Continual Learning for Task-oriented Dialogue Systems}
\label{ch6}

\section{Open Intent Detection \& Learning}
\label{ch6.sec1}

\section{Continual Learning for Semantic Slot Filling}
\label{ch6.sec2}

\section{Continual Learning for Dialogue State Tracking}
\label{ch6.sec3}

\section{Continual Learning for Natural Language Generation}
\label{ch6.sec4}

\section{Joint Continual Learning of all Dialogue Tasks}
\label{ch6.sec5}

\section{Summary}
\label{ch6.sec4}

%% file: ch07.tex
\graphicspath{{./figures/Ch07}}     

\chapter{Continual Learning of Conversational Skills}
\label{ch7}

\section{Learning user behaviors and preferences}
\label{ch7.sec1}

\section{Learning emotions, moods and opinions in dialogues}
\label{ch7.sec2}

\section{Modeling situation-aware conversations}
\label{ch7.sec3}

\section{Summary}
\label{ch7.sec4}

%% file: ch08.tex
\graphicspath{{./figures/Ch08}}     

\chapter{Conclusion and Future Directions}
\label{ch8}